This version of the article has been published in the Journal of Flood Risk Management
DOI: https://doi.org/10.1111/jfr3.70118

# Robust blue-green urban flood risk management optimised with a genetic algorithm for multiple rainstorm return periods

Asid Ur Rehman [1*], Vassilis Glenis [1], Elizabeth Lewis [2], Chris Kilsby [1], Claire Walsh [1]

[1] School of Engineering, Newcastle University, Newcastle Upon Tyne, NE1 7RU, UK

[2] School of Engineering, The University of Manchester, Manchester, M1 7HL, UK

[*] Correspondence: asid-ur-rehman2@newcastle.ac.uk

**Abstract**

Flood risk managers seek to optimise Blue-Green Infrastructure (BGI) designs to maximise return on investment. Current systems often use optimisation algorithms and detailed flood models to maximise benefit-cost ratios for single rainstorm return periods. However, the BGI scheme optimised for one return period (e.g., 100 years) may differ significantly from those optimised for others (e.g., 10 or 20 years). This study aims to assess the effectiveness of single return period-based BGI design across multiple storm magnitudes and introduces a novel multi-objective optimisation framework that simultaneously incorporates five return periods (T = 10, 20, 30, 50, and 100 years). The framework combines a Non-dominated Sorting Genetic Algorithm II (NSGA-II) with a fully distributed hydrodynamic model to optimise the spatial placement and combined size of BGI features. For the first time, direct damage cost (DDC) and expected annual damage (EAD), calculated for various building types, are used as risk objective functions, transforming a many-objective problem into a multi-objective one. Performance metrics such as Median and Maximum Risk Difference (MedRD, MaxRD) between reference and trial Pareto fronts, capturing characteristic single values from the distribution of risk differences, and the Area Under Pareto Front (AUPF), indicating overall optimisation quality, reveal that a 100-year optimised BGI design performs poorly when evaluated for other return periods, particularly shorter ones. In contrast, a BGI design optimised using composite return periods enhances performance metrics across all return periods, with the greatest improvements observed in MedRD (22%) and AUPF (73%) for the 20-year return period, and MaxRD (23%) for the 50-year return period. Furthermore, climate uplift stress testing confirms the robustness of the proposed design to future rainfall extremes. This study advocates a paradigm shift in flood risk management, moving from single maximum to multiple rainstorms-based optimised designs to enhance resilience and adaptability to future climate extremes.

**Keywords**

multi-objective optimisation; genetic algorithm; blue-green infrastructure; multiple return periods; robust flood risk management; climate change resilience

## 1 Introduction

Flooding, a prevalent global natural disaster, exposes 1.47 billion people to 1-in-100-year risk, with annual losses rising from $6 billion in 2005 to an expected $60 billion by 2050 (Salhab & Rentschler, 2020). As urbanisation expands (Miller & Hutchins, 2017; Rentschler et al., 2023) and climate change intensifies, flood events are becoming more frequent and severe (Kendon et al., 2023; Robinson et al., 2021). Traditional urban flood risk management (FRM) approaches, primarily relying on subsurface drainage and designed by historical rainfall data, often struggle to handle extreme and varying rainfall patterns, leaving cities at unprecedented risk (Salinas-Rodriguez et al., 2018). This escalating risk to cities highlights the need for innovative and robust FRM designs to tackle the adverse impact of rainstorm events of varying intensities.

Grey Infrastructure (GI), such as drainage systems and concrete barriers, manages runoff but often fails during severe rainstorms due to limited capacity (D'Ambrosio et al., 2022; Salinas-Rodriguez et al., 2018). Expanding GI can increase capacity, but it is costly, unsustainable, impractical in dense urban areas, and potentially degrades ecosystems (Qin et al., 2013; Rosenbloom, 2018). Such limitations highlight the need for sustainable alternatives such as Blue-Green Infrastructure (BGI) or Low Impact Development (LID). These approaches use semi-natural features such as permeable surfaces, green roofs, rain gardens, and detention/retention ponds to enhance flood risk management by integrating natural hydrological processes like infiltration, evaporation, and temporary storage into urban planning (Alves et al., 2019; O'Donnell et al., 2020; Webber et al., 2020). Beyond managing floods, BGI proves beneficial in boosting biodiversity, mitigating heat islands, and improving urban environments (Ahiablame et al., 2012; Rodriguez et al., 2021). The adaptability, sustainability, and multifunctionality of BGI make it ideal for retrofitting flood management systems by complementing traditional GI methods and driving policy shifts in city councils towards 'blue-green' urban planning (Liberalesso et al., 2020; Manchester City Council, 2021; Wheeler, 2016).

Sustainable FRM requires efficient resource allocation and robust designs that perform across varying rainstorm intensities (Sharma et al., 2021). The literature suggests that the cost-effectiveness and robustness of BGI largely depend on the hydrodynamic models and risk assessment functions used to evaluate design parameters such as BGI type, location, and size.(Maier et al., 2019; Seyedashraf et al., 2021; Wang et al., 2022). For instance, semi-distributed models like the Storm Water Management Model (SWMM) (Rossman, 2010) evaluate BGI efficiency and robustness by considering reduced peak flows and water volumes in drainage networks (Zhi et al., 2022), barely addressing surface flooding and related risks, usually termed stormwater management. Fully distributed models such as CityCAT (Glenis et al., 2018; Iliadis et al., 2023) explicitly simulate BGI to assess its effectiveness in reducing surface runoffs and associated risks to buildings, properties, and infrastructure, providing a more accurate representation of surface FRM. The literature reports that most BGI designs for FRM, developed using detailed flood models, often rely on limited options evaluated through multi-criteria (Alves et al., 2018; Joshi et al., 2021) or scenario-based designs (Abduljaleel & Demissie, 2021; D'Ambrosio et al., 2022; Iliadis et al., 2024; Vercruysse et al., 2019; Webber et al., 2020), which restrict BGI deployment options and may not ensure cost-effective designs. To address this challenge, a multi-objective optimisation algorithm is integrated with a fully distributed hydrodynamic flood model to systematically achieve the most cost-effective BGI design (Ur Rehman et al., 2024). In line with industry practice, a 100-year return period was initially used to optimise the permeable surfaces' location and overall size. However, the 100-year design performed poorly when tried for a 30-year return period, possibly due to a discrete risk objective function, which only accounts for BGI performance in risk reduction when a certain threshold is met. The outcomes highlight the need to investigate BGI optimisation further using continuous risk functions across different return periods. A continuous risk objective should capture the full range of risk reduction, from very minor to maximum. Furthermore, a novel approach is needed to optimise multiple return periods simultaneously, potentially leading to robust FRM. To the authors' knowledge, no such method has yet been developed or tested in multi-objective optimisation, highlighting a considerable research gap.

The prevailing approaches typically focus on a single design storm, often the maximum return period, which may not ensure reliable performance under more frequent or less intense events, raising a fundamental research question of whether this paradigm should be retained or replaced by a more comprehensive approach that integrates multiple rainstorm magnitudes. This study aims to answer this research question by evaluating the effectiveness of a BGI design optimised for a single maximum return period when applied to other return periods, and by developing and comparing a composite BGI design optimised using a range of rainstorm intensities. The novelty of this work lies in the simultaneous incorporation of multiple rainstorm return periods and the design of a continuous risk function for optimisation to develop a comprehensive blue-green flood risk management approach. The specific objectives of this study are to: (i) design a continuous risk objective function for optimisation and analyse subsequent optimisation results for distinct return periods, (ii) incorporate multiple return periods into a multi-objective optimisation framework to achieve a composite BGI design, (iii) quantitatively assess optimisation performance for individual and composite optimised BGI designs, (iv), calculate realistic cost-benefits for BGI lifespan and (v) conduct a stress test to evaluate the climate change resilience of the newly proposed FRM method. The rest of this paper is organised as follows: Section 2 covers the study area and detailed methodology; Section 3 presents the results; and Section 4 discusses the overarching findings, including their implications. Lastly, Section 5 provides the main concluding remarks along with the key limitations and future research directions.

## 2 Material and methods

Figure 1 presents the research flow diagram, with each block representing a distinct stage of the study. Phase I involves BGI and rainstorm design for the selected study area. Phase II uses the outputs from Phase I to develop an optimisation framework by integrating the CityCAT model with an optimisation algorithm. Phase 3 processes the resulting Pareto fronts (a set of optimal solutions), generated for both single and composite return periods, and applies quantitative performance metrics to evaluate optimisation quality. The best-performing BGI design is then carried forward to Phase IV, where a lifetime benefit-cost analysis is conducted alongside stress testing based on projected future rainfall. Sections 2.1 to 2.4 explain each phase in detail.

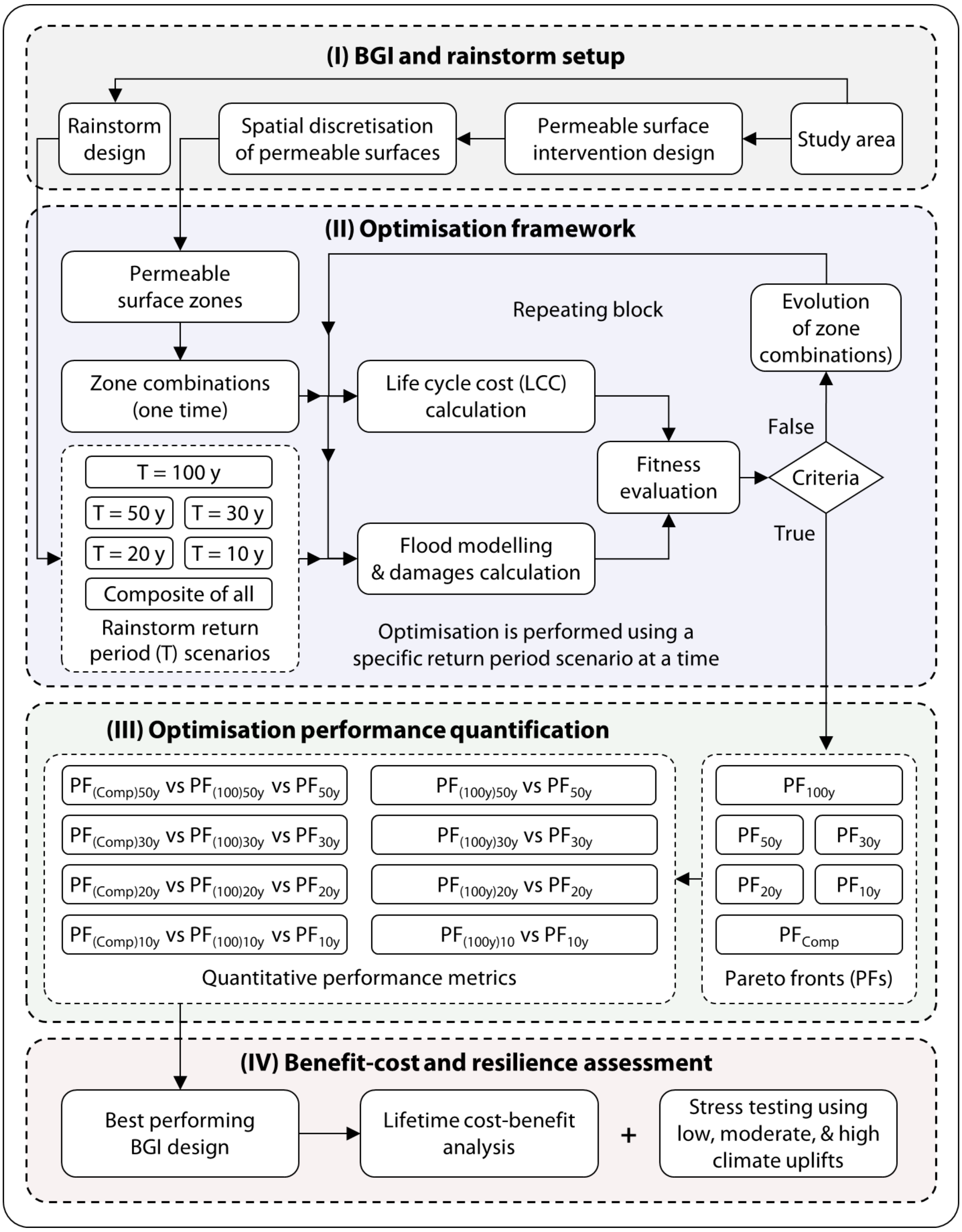

Figure 1. Flow diagram showing the overall research design, including key components, processes, and their interactions. $PF_{50y}$ refers to the Pareto front optimised for a 50-year rainstorm event. $PF_{(100y)50y}$ and $PF_{(Comp)50y}$ denote Pareto fronts obtained by optimising BGI for a 100-year and a composite of multiple return periods rainfall events, respectively and then assessed for performance under a 50-year event. The same annotation applies to other return periods.

### 2.1 BGI and rainstorm setup

#### 2.1.1 Study area

The study area, shown in Figure 2, is an urban catchment in Newcastle upon Tyne. The catchment has a total area of approximately 5.3 km², of which approximately 43% is green/permeable area, 32% is impervious surfaces, and 25% is buildings. Based on type, the building area is further classified into residential (10%) and non-residential (15%) categories. The catchment has a maximum elevation of 128 meters, with a relatively steep slope of 3.3% from northwest to southeast (see the elevation map in *Supplementary Information S2*). The high gradient allows rainstorm water to move relatively quickly from the upper catchment to the lower catchment, placing infrastructure and properties in the lower catchment at high risk.

Newcastle has experienced numerous flash floods due to heavy rainstorm events (Newcastle City Council, 2016), the most severe being 'Thunder Thursday', during which the city received 49 mm of rainfall in 2 hours (Environment Agency, 2012). This rainstorm was approximated as a 100-year return period event.

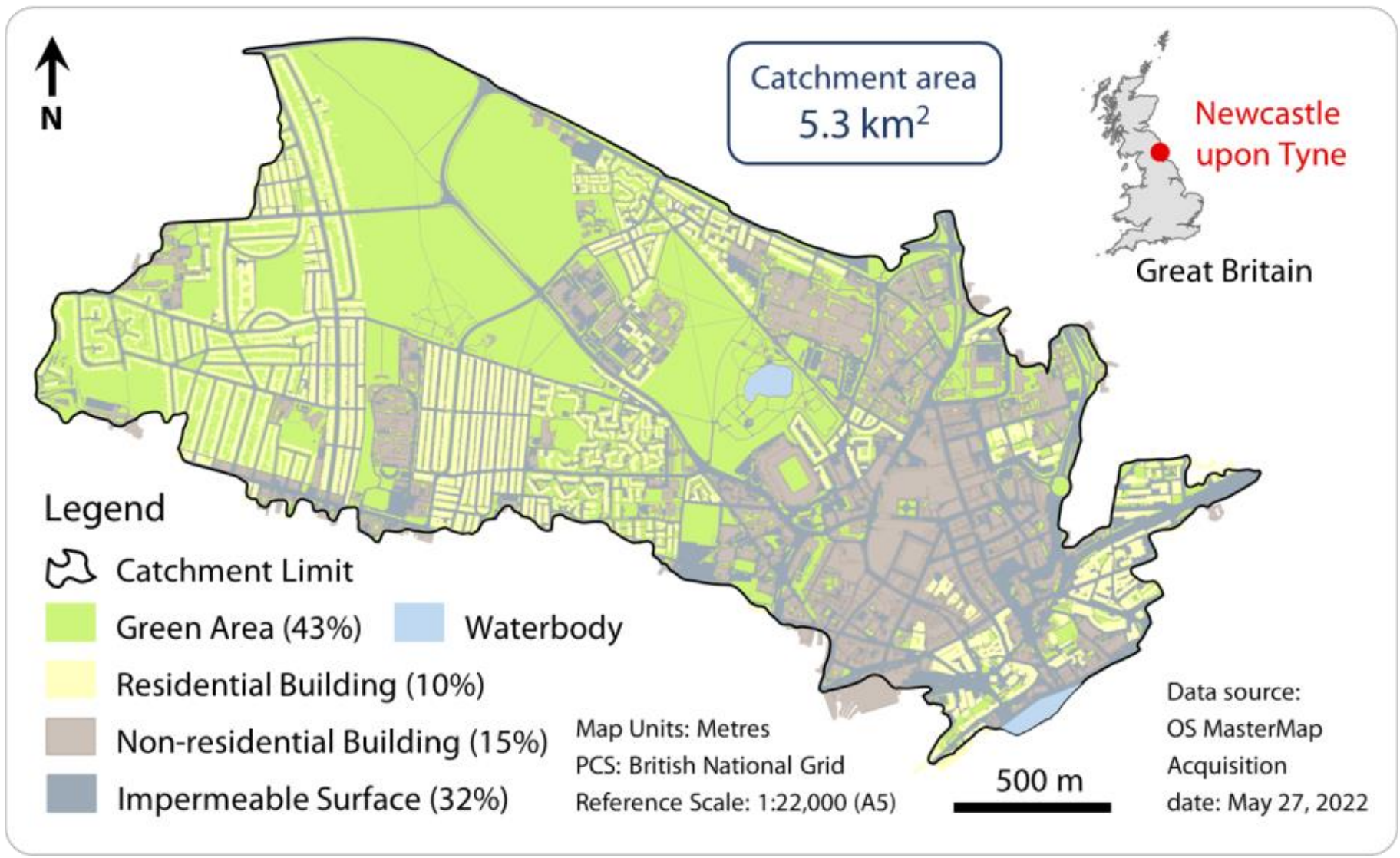

Figure 2. Map of the study area

#### 2.1.2 Permeable surface intervention design

The current study builds on previous research (Ur Rehman et al., 2024), which assessed the cost-effectiveness of permeable surface interventions by dividing the entire catchment into zones of varying sizes and quantities. The best-performing scenario, featuring 80 permeable zones (shown in Figure 3), was selected for this case. The total permeable surface area is 0.74 km², comprising 31% parking areas and 69% roadside pavements and paths. As shown on the map, the sizes of permeable surface zones vary, but the optimisation process automatically normalises these differences through intervention costs.

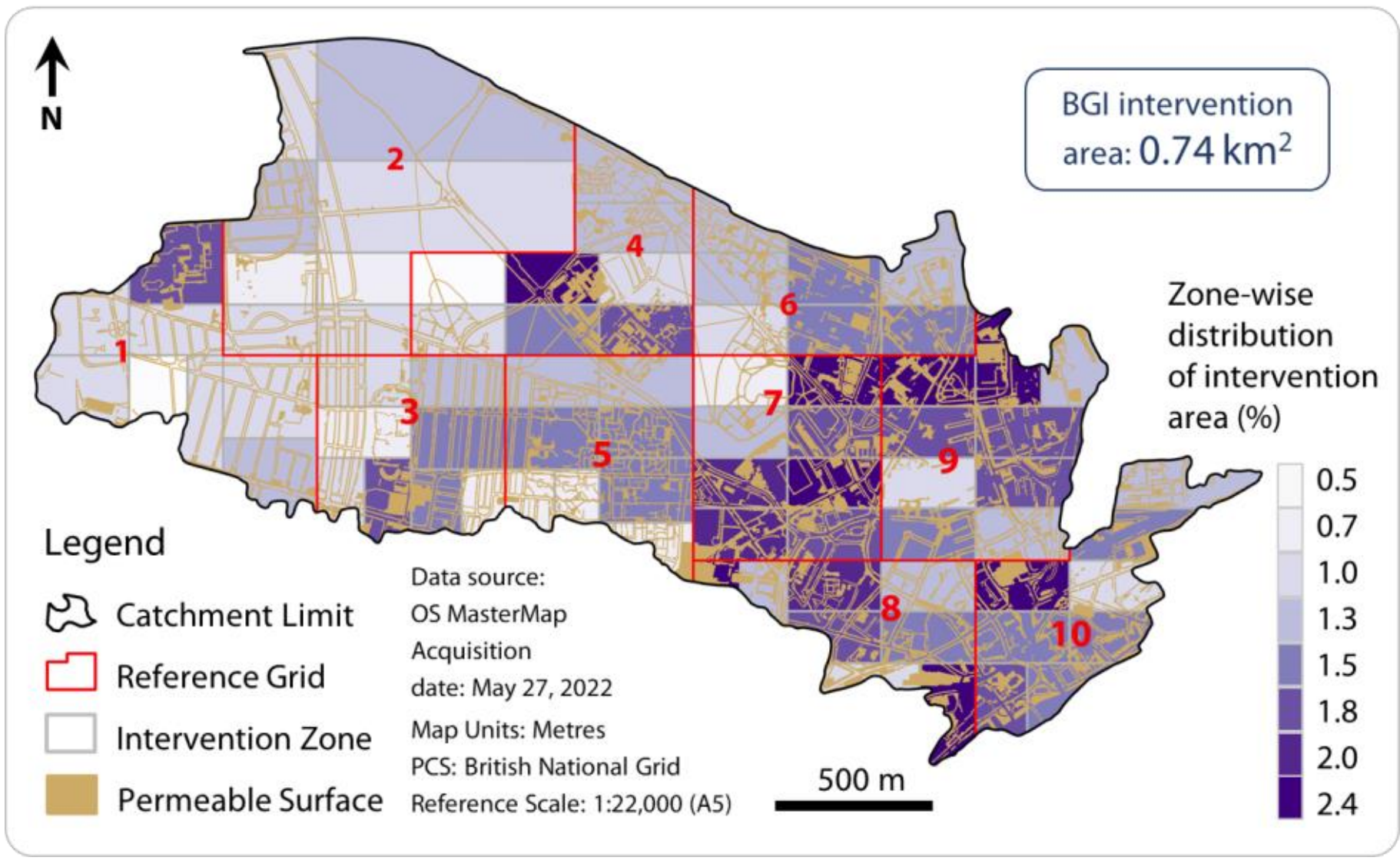

Figure 3. Permeable surface zonation for interventions: A total of 80 zones were defined, with red numbered boxes used to reference different parts of the catchment.

### 2.1.3 Rainstorm design

The standard Depth-Duration-Frequency (DDF) model for the UK (Faulkner, 1999) was used to calculate the total rainfall amount for 100, 50, 30, 20, and 10-year return periods (T), each with a duration (t) of 30 minutes. The DDF model equation used for rainfall with a duration of 12 hours or less is given below:

$$ln(R) = (cy + d_1)\, ln(D) + er + f \quad where\ r = -ln\left[-ln\left(1 - \frac{1}{T}\right)\right]$$

R is the rainfall depth, D is the duration, r is the Gumbel reduced variate, T is the return period, and c, $d_1$, e, and f are catchment descriptors. The hyetograph generation method from the Flood Studies Report (FSR) (Institute of Hydrology, 1975) was then applied to generate temporal distribution profiles for the given rainfall events in an urban catchment. The equation to create the rainfall distribution profile is given below:

$$y = \frac{1 - a^z}{1 - a}\ where\ z = x^b$$

Where: y is the fraction of the rainfall that drops within the proportion x of the total rainstorm duration. Parameters a and b have fixed values, and the profile is centred on the peak. The obtained rainfall totals and temporal distributions for the considered return periods with the same durations are shown in Figure 4.

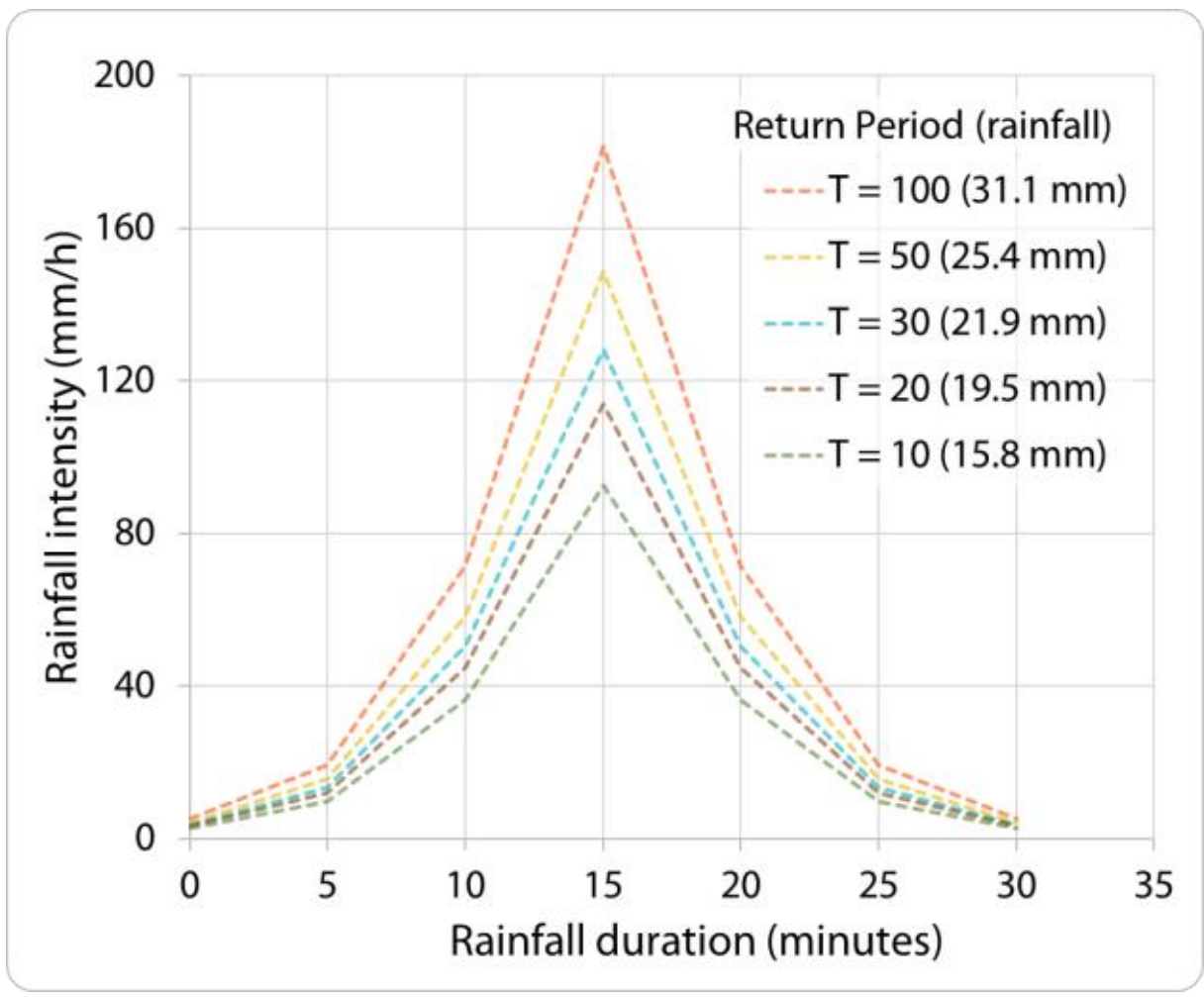

Figure 4. Design rainstorms for five return periods

## 2.2 Optimisation framework

The optimisation framework integrates a Multi-Objective Optimisation Algorithm (MOOA) (Venter, 2010), called the Non-dominated Sorting Genetic Algorithm (NSGA-II) (Deb et al., 2002), with the CityCAT flood simulation model to identify the most cost-effective spatial combinations of permeable surface zones. MOOAs generally solve optimisation problems by iteratively changing the values of decision variables in a set of candidate solutions, called a population, to achieve better values for the considered objective functions. This iterative process, known as generations, continues until a termination criterion is reached. The result is a set of optimal solutions called the Pareto optimal front, representing the best trade-offs between objective functions. NSGA-II, in particular, uses non-dominated sorting to identify the optimal solutions and applies a crowding distance method to maintain population diversity to create better solutions in the new generation (Deb et al., 2002). For a detailed overview of a genetic algorithm, readers are encouraged to refer to the article by Maier et al. (2019). NSGA-II was selected over other optimisation algorithms due to its straightforward implementation and its ability to avoid trapping in local optima, thereby achieving a good balance between adaptability and optimisation efficiency. CityCAT was chosen for hydrodynamic modelling because of its advanced capabilities in accurately representing and modelling buildings, BGI, and other land use features (Iliadis et al., 2023), which enables precise calculation of water depths and velocities. While Sections 2.2.1 to 2.2.4 offer an overview of these topics, Ur Rehman et al. (2024) provide a thorough justification for prioritising NSGA-II, detailed insights into CityCAT's functionality, and a step-by-step guide to integrating both tools within the context of flood risk optimisation.

The following sub-sections provide a concise overview of hydrodynamic modelling and the optimisation processes implemented in this study.

### 2.2.1 Hydrodynamic flood modelling

This study uses the advanced City Catchment Analysis Tool (CityCAT) (Glenis et al., 2018). CityCAT is a fully distributed, physically-based hydrodynamic model that uses a high spatial resolution Digital Elevation Model (DEM) to explicitly represent land use features, including buildings, permeable green spaces, and impervious areas. It also simulates the hydrodynamics of BGI, incorporating features such as permeable surfaces, green roofs, and water butts. Surface water elements like detention ponds and rivers are represented within the DEM. CityCAT takes rainfall, DEM, land use and BGI inputs and applies shallow water equations to simulate 2D water depths and velocities across the spatial domain. It can also integrate subsurface sewer networks to compute coupled surface-subsurface water flows and volumes, though this significantly increases computational time. Due to such computational constraints, this study only uses CityCAT's 2D surface flood simulation module. Further details on its role within the optimisation framework are provided in Section 2.2.3. For a comprehensive overview of CityCAT, refer to Ur Rehman et al. (2024), with detailed information in Glenis et al. (2018). A detailed description of the data used for BGI and flood modelling is provided in *Supplementary Information S1*.

CityCAT has been successfully validated in various studies using laboratory datasets (Glenis et al., 2018), on-the-ground flood depths derived from social media images (Kutija et al., 2014), and property-level flooding based on local authority surveys (Bertsch et al., 2022). More recently, Iliadis et al. (2023) validated CityCAT outputs for the 2012 "Thunder Thursday" flood event and found good agreement between the model and surveyed data. Notably, the validation area overlaps with the catchment considered in the present study. Therefore, the same CityCAT parameters were adopted for this study (see *Table S3 in Supplementary Information S1*).

### 2.2.2 Decision variables and objective functions

**The decision variables** in this optimisation problem correspond to the spatial locations of 80 predefined permeable surface zones. Each candidate solution is encoded as a binary vector of length 80, where a value of 1 indicates the inclusion of a particular zone, and 0 indicates its exclusion. With 80 binary decision variables, the total solution space consists of $2^{80}$ possible combinations, which is an extremely large number. However, the binary representation allows the algorithm to efficiently explore a limited range of spatial configurations to find the optimal set of solutions. The encoding strategy used in this study is described in detail by Ur Rehman et al. (2024).

**The objective functions** are the life cycle cost (LCC) of the permeable surface zones and the associated levels of risk (R) to properties. Based on these decision variables and objective functions (LCC, R), the optimisation problem can be written as:

$$Minmise: F(I) = (F_{LCC}, F_R)$$

The LCC for the $j^{th}$ permeable zone can be computed using the following equation:

$$LCC(I_j) = (C_c + C_o) \times S(I_j) \quad j = 1, \ldots, 80$$

$C_c$ and $C_o$ are the unit size capital and operational costs, respectively, and $S(I_j)$ is the $j^{th}$ permeable surface zone area. The capital cost refers to the one-time installation expense, while the operational cost is the annual maintenance expense of the permeable surfaces over a specific lifetime.

The LCC per unit of permeable surface area was calculated following the guidelines from the UK Environment Agency (Gordon-Walker et al., 2007). LCC comprises one-time Capital Expenditures (CapEx) and recurring Operational Expenditures (OpEx) over the BGI lifetime. A lifespan of 40 years was assumed for permeable pavements. An average inflation rate of 2.9%, calculated from the UK Office for National Statistics' Consumer Price Index (CPI) spanning 2007 to 2022, was applied to calculate the current and future costs using the equation below.

$$FV = BV(1 + i)^n$$

Where: FV represents the future value, BV is the base year value, i is the inflation rate, and n is the number of maintenance years. For this study, the operational cost was multiplied by the lifespan of the permeable surface, which is 40 years, before being added to the capital cost. The calculated CapEx was £82.9 $m^{-2}$, with an OpEx of £46.6 $m^{-2}$, resulting in an LCC of £129.5 $m^{-2}$.

**The risk objective function** ($F_R$) in this study represents two scenarios: (i) a risk function based on a single return period and (ii) a composite risk function incorporating multiple return periods. According to the Intergovernmental Panel on Climate Change (IPCC) Sixth Assessment Report, flood risk results from the combination of hazard, exposure, and vulnerability of affected systems (IPCC, 2023):

$$Risk_{IPCC} = Hazard \times \text{Exposure} \times \text{Vulnerability}$$

Previously, Ur Rehman et al. (2024) approximated flood hazard by extracting flood depths using hydrodynamic simulations. These depths were used as inputs into a building-level exposure tool (Bertsch et al., 2022) to estimate the number of buildings exposed to flooding. The resulting building count served as a proxy for flood risk within the objective function. However, this approach presents two key limitations. First, it omits building vulnerability, thereby failing to capture the true nature of flood risk. Second, the exposure tool produces discrete outputs, classifying buildings as "exposed" only when both the 90$^{th}$ percentile and the mean of maximum flood depths exceed predefined thresholds. This binary classification leads to a discrete risk function, which only recognises BGI effectiveness when these thresholds are surpassed. Consequently, it may underestimate the efficiency of BGI interventions and produce inconsistent performance across varying storm intensities.

To address these limitations, this study adopts a revised approach by adapting the risk estimation methodology outlined in the Multi-Coloured Manual (MCM) for the Economic Appraisal of Flood Risk (Penning-Rowsell et al., 2014). The MCM defines flood risk as:

$$Risk_{MCM} = \text{Probability} \times \text{Consequence}$$

When aligned with the IPCC framework, probability in the MCM corresponds to hazard, while consequence encompasses both exposure and vulnerability. Although the MCM does not explicitly define vulnerability, it is implicitly embedded within the consequence term, which is calculated as:

$$\text{Consequence} = \sum Flood\ depth \times Property\ type \times Damage\ factor$$

In this formulation, flood depth and property type represent exposure, whereas the damage factor is linked to building vulnerability. The consequence term is operationalised through depth-damage curves, which are derived from empirical data and express the percentage of damage relative to the asset value across varying flood depths, as illustrated in Figure 5.

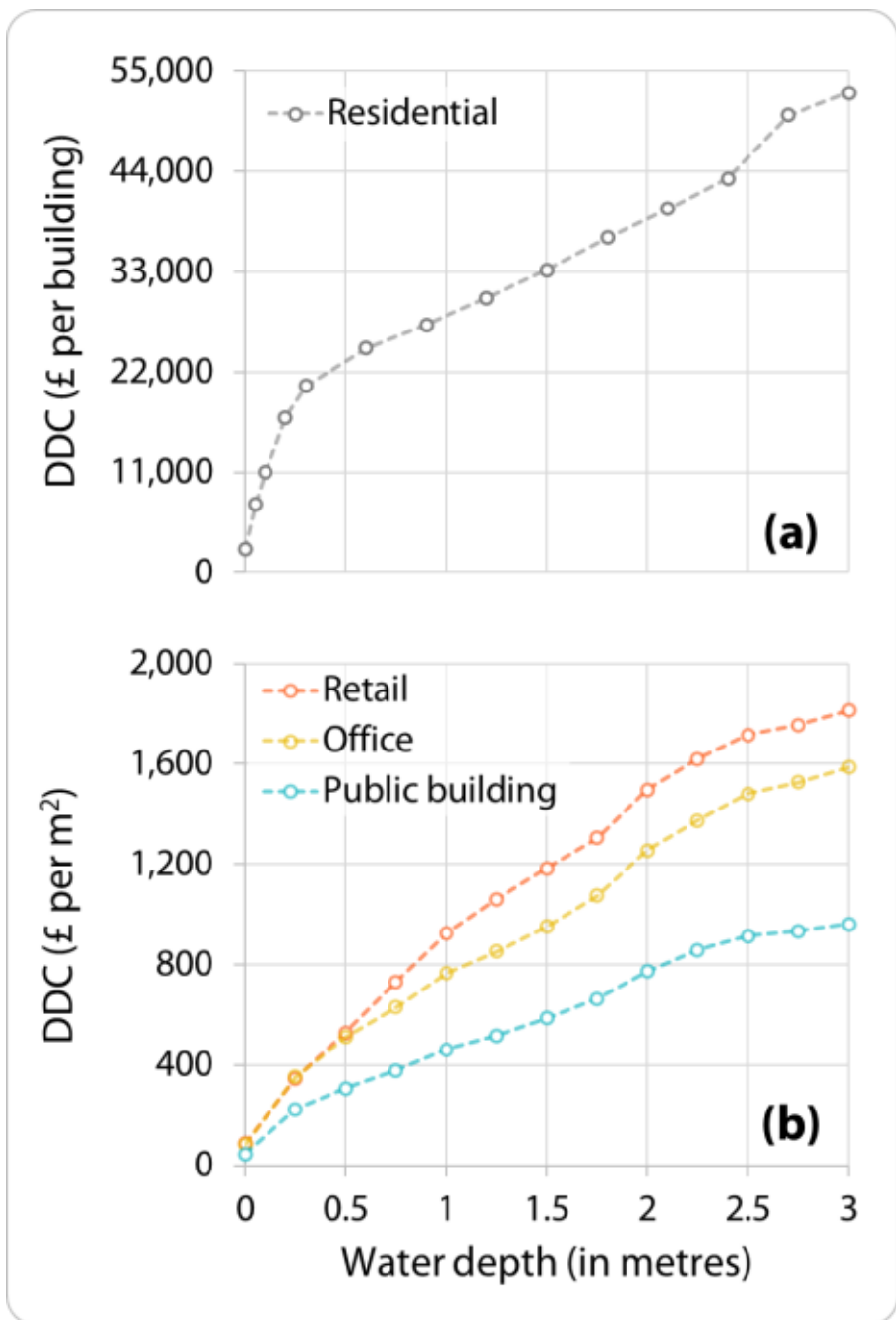

Figure 5. Depth damage curves for (a) residential and (b) non-residential buildings (Source: Penning-Rowsell et al., 2014).

The adaptation of depth-damage curves enables a continuous and more representative estimation of flood risk, capturing the varying susceptibility of different building types to damage across a spectrum of flood depths (i.e. hazard severity). Using these curves, this study calculates Direct Damage Cost (DDC) and Expected Annual Damage (EAD) as risk objective functions for optimisation under single and multiple return periods.

According to the MCM, DDC is calculated per property for residential buildings (Figure 5a) and per unit area for non-residential buildings (Figure 5b), with the latter multiplied by the total floor area to obtain total damage to the building (see Section 2.2.3). EAD is then computed by integrating DDC values across different return periods (see Section 2.2.4 for details). While DDC represents only the consequence component of risk, EAD accounts for both probability and consequence, offering a more comprehensive measure of flood risk.

### 2.2.3 Single return period-based BGI optimisation

The schematic diagram of the multi-objective optimisation framework is illustrated in Figure 6. It comprises NSGA-II, CityCAT, and Risk Modules. Initially, NSGA-II (Figure 6a) randomly generates 'P-2' candidate solutions, each representing different combinations of permeable surface zones. The first candidate solution, with no permeable zones, defines the baseline scenario, while the $P^{th}$ candidate solution, with all permeable zones, represents the maximum intervention scenario. Next, NSGA-II passes these candidate solutions to the objective functions for fitness evaluation. The cost function calculates the LCC of each candidate solution using the formula provided below:

$$LCC(S_z) = \sum_{j=1}^{80} LCC(I_j) \quad \begin{matrix} where \ z = 1,2,3....P \\ I_j \neq 0 \end{matrix}$$

$LCC(S_z)$ is the life cycle cost for the $z^{th}$ candidate solution, and $LCC(I_j)$ is the life cycle cost of the $j^{th}$ permeable intervention zone.

Similarly, the risk function processes the 'P' candidate solutions sequentially using the CityCAT module (Figure 6b). This module consolidates geometry files of the available permeable surface zones in each solution. Then, it incorporates standard inputs such as rainfall, DEM, and geometries of green areas and building footprints to compute maximum water depths. These water depths are then forwarded to the risk module (Figure 6c), which is adapted from Bertsch et al. (2022). The risk module creates a buffer around each building at 150% of the DEM grid cell dimension. The buffer polygons are overlaid on the flood depth map to extract flood depth cells surrounding each building. The risk module then calculates the $90^{th}$ percentile and mean of maximum depth values within each building buffer area to determine whether the building $B_i$ (where i = 1, 2, …, m) is at risk or not at risk. If $B_i$ is categorised as at risk, the DDC is calculated using the $90^{th}$ percentile of the water depth value, which helps tackle outliers in maximum flood depth values. The building classification approach helps avoid extra DDC for buildings with negligible water depths. The total risk is the sum of DDCs for 'm' buildings, calculated using the following equations:

$$DDC(S_z) = \sum_{i=1}^{m} \begin{cases} I_{R_i} \times DDC_{res_d_{90th_i}} & B_i \text{ is residential} \\ I_{R_i} \times A_{B_i} \times DDC_{non-res_d_{90th_i}} & B_i \text{ is non} - \text{residential} \end{cases} \quad z = 1,2,3, \ldots P \tag{1}$$

$$I_{R_i} = f\left(d_{m_i}, d_{90th_i}\right) = \begin{cases} 0 & d_{m_i} < 0.1\, m \;\; AND \;\; d_{90th_i} < 0.3\, m & B_i \text{ not at risk} \\ 1 & Otherwise & B_i \text{ at risk} \end{cases} \tag{2}$$

Where:

- $\boldsymbol{DDC(S_z)}$ is the direct damage cost (risk level) for the z<sup>th</sup> candidate solution.
- $\boldsymbol{I_{R_i}}$ is the risk index, a function of the mean maximum depth $(\boldsymbol{d_{m_i}})$ and the 90th percentile of maximum depths ($\boldsymbol{d_{90th_i}}$) around the i th building $\boldsymbol{B_i}$. The criterion for $\boldsymbol{I_{R_i}}$ is adapted from the exposure classification scheme presented by Bertsch et al. (2022).
- $\boldsymbol{DDC_{res_d_{90th_i}}}$ is the direct damage cost for a residential building based on the 90th percentile of maximum flood depths around $\boldsymbol{B_i}$.
- $\boldsymbol{A_{B_i}}$ is the area of $\boldsymbol{B_i}$, and $\boldsymbol{DDC_{non\text{-}res_d_{90th_i}}}$ is the direct damage cost per unit area for a non-residential building, depending upon the 90th percentile of maximum flood depths around $\boldsymbol{B_i}$.

After receiving the LCCs and DDCs for ‘P’ candidate solutions, the NSGA-II module evaluates their fitness based on their ability to minimise both LCC and DDC simultaneously. The algorithm then checks the generation number as a termination criterion and stops if it is met. Otherwise, evolutionary operations, including parent selection, crossover, and mutation (Ur Rehman et al., 2024), create a new set of ‘P’ offspring. These offspring are evaluated for cost and risk objectives to determine their LCCs and DDCs. NSGA-II combines the LCCs and DDCs of both parent and offspring populations, selects the ‘P’ best solutions, and forms a new generation of parents. This process repeats for N generations to achieve optimal solutions with the lowest LCC and DDC.

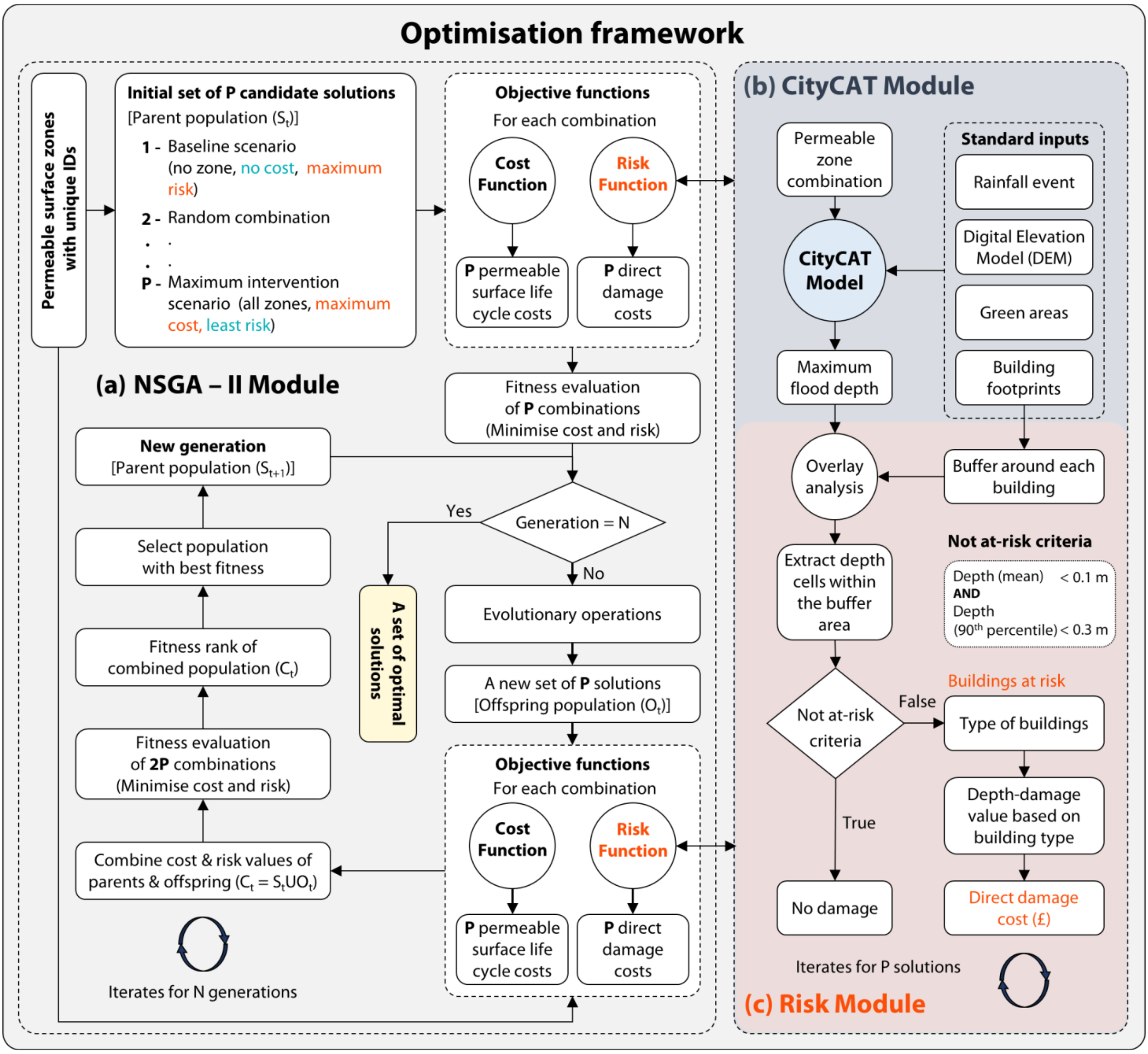

Figure 6. The optimisation framework for individual return periods. The NSGA-II module (a) generates a population of candidate solutions, calculates the life cycle cost (LCC) for each solution, and forwards them one by one to the CityCAT module (b), which simulates each solution to determine maximum flood depths. These depths are then used by the Risk module (c) to identify the number of buildings at risk and estimate the associated direct damage cost (DDC). The total damage cost for each solution is calculated and returned to the NSGA-II module, which evaluates LCC versus DDC to identify better-performing solutions and evolves them to create a new population. This process continues until the termination criterion is met, ultimately providing the optimal solutions.

### 2.2.4 Multiple return period-based BGI optimisation

For multiple return periods, each return period has an associated direct damage cost (DDC), resulting in multiple DDC objectives. Traditional multi-objective algorithms are typically limited to handling up to three objectives efficiently. A potential alternative is a many-objective optimisation algorithm (Deb & Jain, 2014), which extends multi-objective methods to accommodate and tackle a larger number of objectives. However, the implementation of such algorithms is complex, and, in this case, assigning equal weight to DDCs for frequent and rare events does not reflect real-world risk priorities, making this approach unsuitable for the current study. Instead, a weighted aggregation of DDCs into a single risk objective function, such as expected annual damage (EAD), is more appropriate. EAD, calculated using the trapezoidal rule, assigns weights to respective DDCs based on their exceedance probability (Bilskie et al., 2022). Five return periods (T = 10, 20, 30, 50, and 100 years) were selected to balance accuracy and computational cost, as Ward et al. (2011) noted that using very few return periods may overestimate EAD. Adopting EAD as a risk objective function provides a more accurate representation of risk and enables the calculation of realistic lifetime cost benefits for BGI. The following equation from the Scottish Government (2018) was used to compute EAD:

$$EAD(S_z) = \frac{1}{2}\begin{bmatrix} (DDC(S_z)_{T10} + DDC(S_z)_{T20}) \times \left(\frac{1}{10} - \frac{1}{20}\right) + \\ (DDC(S_z)_{T20} + DDC(S_z)_{T30}) \times \left(\frac{1}{20} - \frac{1}{30}\right) + \\ (DDC(S_z)_{T30} + DDC(S_z)_{T50}) \times \left(\frac{1}{30} - \frac{1}{50}\right) + \\ (DDC(S_z)_{T50} + DDC(S_z)_{T100}) \times \left(\frac{1}{50} - \frac{1}{100}\right) + \\ (DDC(S_z)_{T100} + D(S_z)_{INFIN}) \times \left(\frac{1}{100} - 0\right) \end{bmatrix} where\ z = 1,2, \ldots P \quad (3)$$

$$D(S_z)_{INFIN} = DDC(S_z)_{T100} + (DDC(S_z)_{T100} - DDC(S_z)_{T50}) \times \left(\left(\frac{1}{100} - 0\right) \Big/ \left(\frac{1}{50} - \frac{1}{100}\right)\right) \quad (4)$$

The term $D(S_z)_{INFIN}$ is used to get proportionate direct damage contributions from the 100-year return period in the absence of the next longer return period.

The adaptation of a multi-objective optimisation framework for composite return periods is shown in Figure 7. While the processes in the NSGA-II module remain unchanged, the CityCAT and risk modules evaluate each candidate solution across all return periods, calculating individual DDCs and then the EAD for each solution. This process is repeated ‘P’ times to generate ‘P’ EADs. The NSGA-II module then optimises over N generations using the LCC and EAD values of the candidate solutions.

*Table S4 in Supplementary Information S4* provides the NSGA-II parameters and values used for single- and multiple-return-period optimisation.

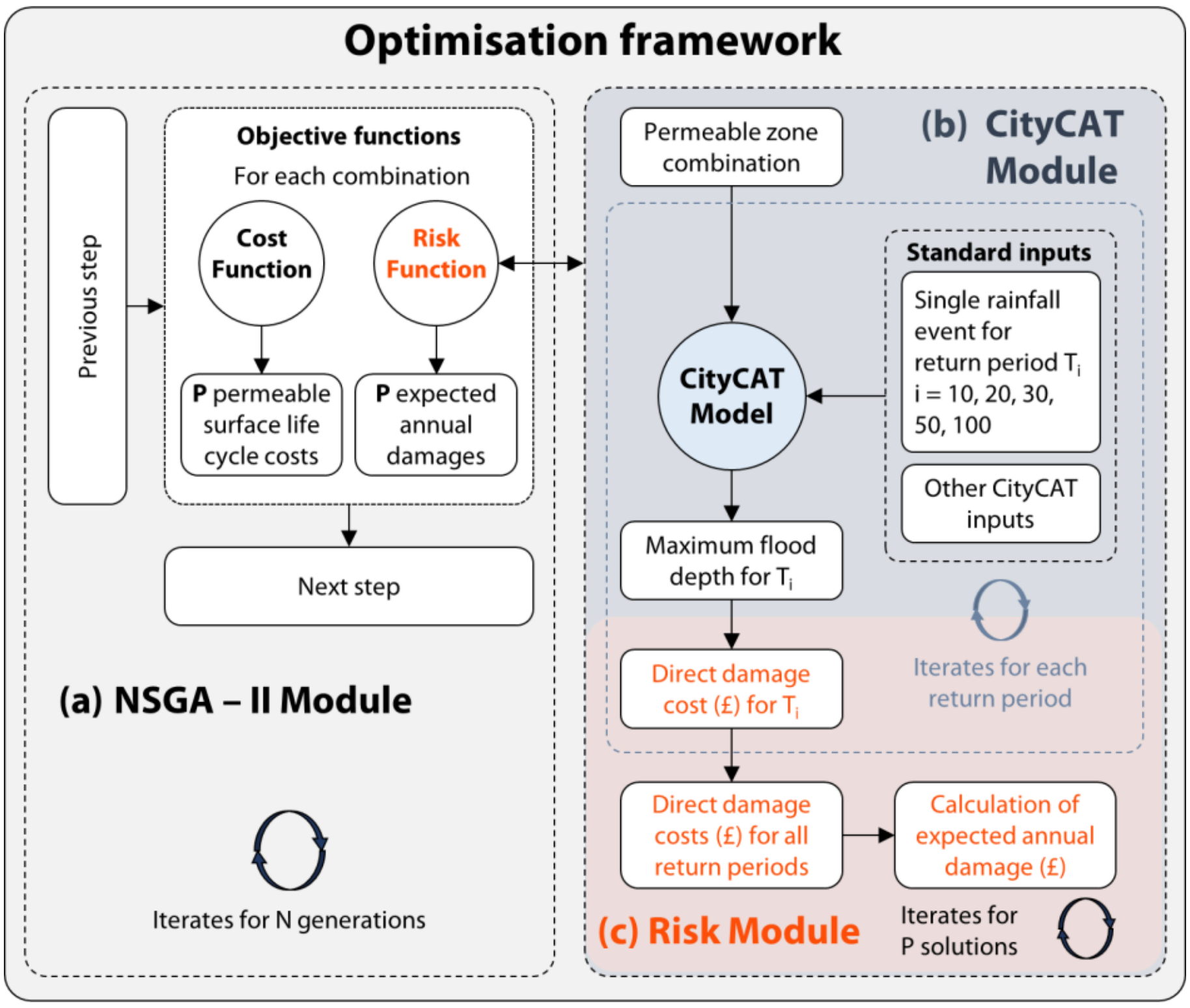

Figure 7. The optimisation framework for the composite of multiple return periods. The optimisation process functions similarly to that described in Figure 6. The key difference is that instead of using a single direct damage cost (DDC) as the risk function, the Risk module calculates the expected annual damage (EAD) by combining DDC values across multiple return periods. The NSGA-II module then evaluates life cycle cost (LCC) against EAD.

## 2.3 Optimisation performance quantification

Three metrics, Maximum Risk Difference (MaxRD), Median Risk Difference (MedRD), and Area Under Pareto Front (AUPF), were developed to quantify the performance and robustness of Pareto fronts obtained for the 100-year and composite return period optimisation (trialled Pareto fronts). Pareto fronts obtained by optimising BGI for individual return periods were used as a reference.

**Maximum Risk Difference (MaxRD) and Median Risk Difference (MedRD)** quantify the largest and median disparity in risk (DDC) between the reference and trialled Pareto fronts at any BGI cost (LCC). MedRD represents the central tendency of the risk discrepancies, providing a robust measure less sensitive to extreme deviations than the maximum difference. The formulas for these metrics are provided below:

$$MaxRD = \max_{c \in C} \left| R_{trial}(c) - R_{ref}(c) \right| \tag{5}$$

$$MedRD = median\left( \left| R_{trial}(c_1) - R_{ref}(c_1) \right|, \left| R_{trial}(c_2) - R_{ref}(c_2) \right|, \ldots, \left| R_{trial}(c_n) - R_{ref}(c_n) \right| \right) \tag{6}$$

Where:

$R_{ref}$ (c) and $R_{trial}$ (c) represent the risk value (DDC) on the reference and trialled Pareto fronts for a given cost c. C = {$c_1$, $c_2$, . . . , $c_n$} denotes the set of all cost values (LCC) where the solutions of Pareto fronts are evaluated.

MaxRD and MedRD metrics can also be expressed as percentages of the total risk range to get normalised values. The total risk range ($R_{range}$) is defined as the difference between the risk of the baseline or no intervention solution ($DDC_{Baseline}$) and the maximum intervention solution with the highest LCC ($DDC_{Max\text{-}int}$).

$$MaxRD\ (\%) = \frac{MaxRD}{R_{range}}\ X\ 100 \tag{7}$$

$$MedRD\ (\%) = \frac{MedRD}{R_{range}}\ X\ 100 \tag{8}$$

$$R_{range} = DDC_{Baseline} - DDC_{Max-int} \tag{9}$$

**Area Under Pareto Front (AUPF)** measures the total area beneath a Pareto front in the objective space, defined by the BGI cost (LCC) and risk (DDC). This metric captures the performance of the entire Pareto front distribution. A smaller area typically indicates a more optimal front, representing lower cost and risk. AUPF is computed using the trapezoidal rule, as defined in the equation below:

$$AUPF = \sum_{i=1}^{n-1} \frac{(c_{i+1} - c_i)\ X\ (R(c_i) + R(c_{i+1}))}{2} \tag{10}$$

Where:

C= {$c_1$, $c_2$, . . . , $c_n$} represents the set of sorted cost values (LCC) and $R(c_i)$ = risk value (DDC) at cost $c_i$.

The difference in underneath areas (ΔAUPF) between the reference ($AUPF_{ref}$) and trialled Pareto fronts ($AUPF_{trial}$) reflects the overall disparity in the trialled set of solutions. This disparity can be expressed as a percentage of the reference AUPF to get normalised values.

$$\Delta AUPF\ (\%) = \frac{\Delta AUPF}{AUPF_{ref}}\ X\ 100 \tag{11}$$

$$\Delta AUPF = AUPF_{trial} - AUPF_{ref} \tag{12}$$

**Interpretation of performance evaluation metrics** requires analysing their values in synergy. Percent MaxRD and MedRD capture characteristic single values from the distribution of risk differences and thus represent local improvements. In contrast, the percent ΔAUPF quantifies the overall gap in trialled solutions, representing global improvements. Referring to Table 1, it is also important to note that ΔAUPF (%) has an unbounded upper limit; therefore, it should not be interpreted in isolation and must be considered alongside other metrics to enable meaningful comparisons. To achieve optimal performance, all three metrics should ideally be as close as possible to their minimum values. A combination of low MedRD (%) and ΔAUPF (%) with a high MaxRD (%) suggests overall good optimisation performance with a few poorly performing trialled solutions. Conversely, moderate ΔAUPF (%) alongside high MedRD (%) and MaxRD (%) indicates inconsistent optimisation, where some trialled solutions are highly effective but majority may differ significantly from the reference front. High values for all three metrics imply that the trialled solutions are totally incompatible with the reference set.

| No. | Performance metric | Lower bound value (%) | Upper bound value (%) |
|---|---|---|---|
| 1 | MaxRD (%) | 0 | 100 |
| 2 | Med RD (%) | 0 | 100 |
| 3 | ΔAUPF (%) | 0 | Unbounded (depends on deviation severity) |

Table 1. Upper and lower bounds of performance metrics

## 2.4 Benefit-cost and resilience assessment

### 2.4.1 Benefit-cost analysis of composite optimised BGI design

Integrating BGI LCC and EAD in multiple-return periods-based optimisation provides an opportunity to calculate more realistic benefit-cost ratios by accounting for the lifespan of the BGI. The following equation was used to compute benefit-cost ratios for optimised solutions:

$$B/_C = \frac{(EAD_{Baseline} - EAD_{BGI})\ X\ BGI\ lifespan}{BGI\ LCC} \tag{13}$$

Where:

- B/C represents the benefit-cost ratio.
- BGI lifespan is the time (in years) when substantial maintenance or overhauling of the BGI is required.
- BGI LCC is the life cycle cost of the BGI, calculated based on its lifespan.
- $EAD_{Baseline}$ is the expected annual damage for the baseline flooding scenario (i.e., without BGI intervention).
- $EAD_{BGI}$ is the expected annual damage after implementing a specific BGI solution.

### 2.4.2 BGI design resilience assessment

A stress test using three climate change uplifts, classified as low, medium, and high, assessed the resilience of the proposed FRM design. Based on Chan et al. (2023), rainfall increases of 15%, 30%, and 45% were applied to baseline return periods for each uplift. Figure 8 shows how these uplifts can be considered equivalent to an increase in the return period. For example, a 45% rainfall increase for a 100-year storm equates to a 360-year return period on the baseline scale.

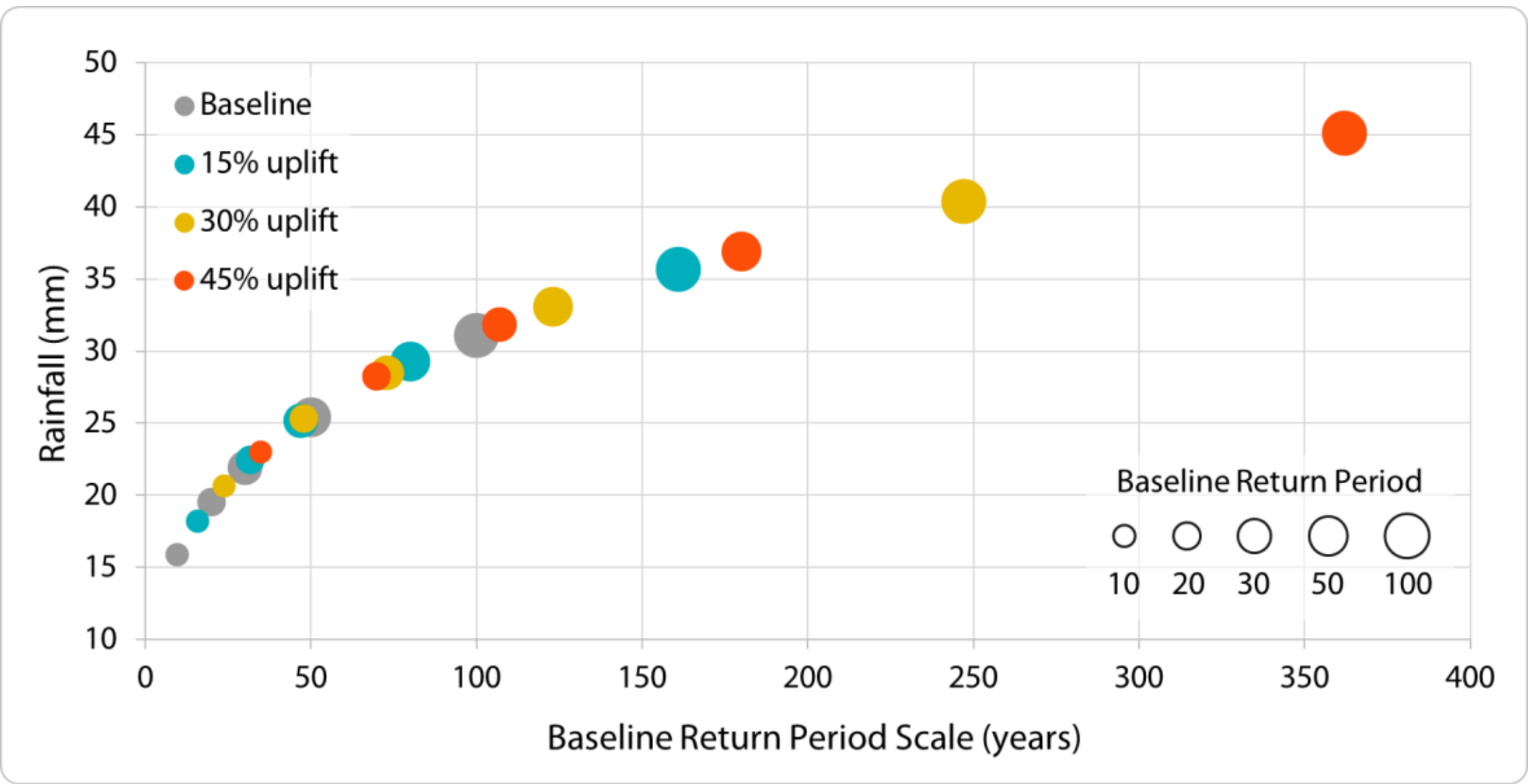

Figure 8. Different climate uplifts drawn on baseline return periods (percent uplifts were adapted from Chan et al. (2023))

## 3 Results

### 3.1 Optimisation for individual return periods

The BGI optimisation results for individual return periods are shown in Figure 9a-e. Referring to the scatter plots, the x- and y-axes represent the BGI life cycle cost (LCC) and the buildings' direct damage cost (DDC), respectively. Grey dots in the plots depict solutions generated throughout the evolutionary process of NSGA-II, whereas the coloured dots indicate the optimal or best solutions achieved by the final generation. The optimal solutions for each return period form a curve, representing the Pareto optimal front. On each Pareto front, the solution positioned at the top-left represents the baseline scenario, while the bottom-right solution corresponds to the maximum intervention scenario. The latter indicates that the solution has the most permeable surface zones with the highest LCC. Additionally, for each return period, an individual spatial map highlights the contribution of each permeable surface zone to the optimal solutions. Zones with the highest contribution, indicated in dark blue, are considered highly cost-effective, while zones in orange shades are the least cost-effective. Zones with no contribution to risk reduction are shown in white.

Figure 9a-e highlights BGI's performance in reducing DDC and identifying the best zones for permeable interventions. The Pareto curve for the 50-year return period achieves greater depth than any other return period, indicating that optimal solutions with smaller LCCs (up to ~£10 million) result in a higher reduction in total DDC than other return periods with similar LCCs. In contrast, the Pareto curve for the 30-year return period has relatively low curvature towards the minimal value.

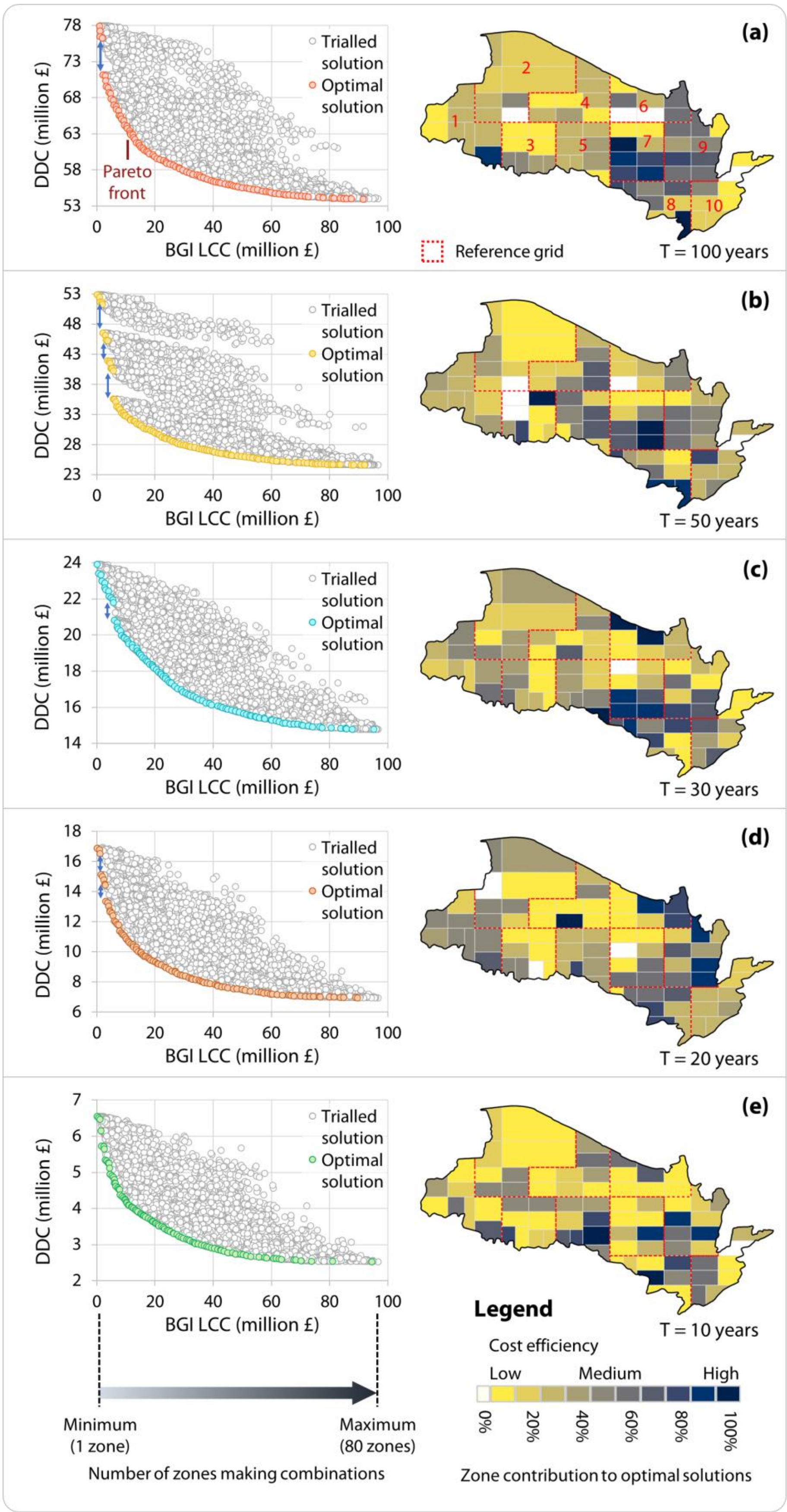

Figure 9. Optimisation of permeable zones for (1) 100-year, (b) 50-year, (c) 30-year, (d) 20-year, and (e) 10-year rainstorm return periods. Each map illustrates the contribution (in percent) of each zone to the number of solutions in the respective Pareto front, with higher contributions indicating greater cost-effectiveness. Grey-outlined points represent all solutions evaluated during the optimisation process, while coloured points denote the optimised solutions for the respective return period. Tinny blue arrows along the y-axis indicate abrupt gaps between consecutive solutions along the Pareto fronts.

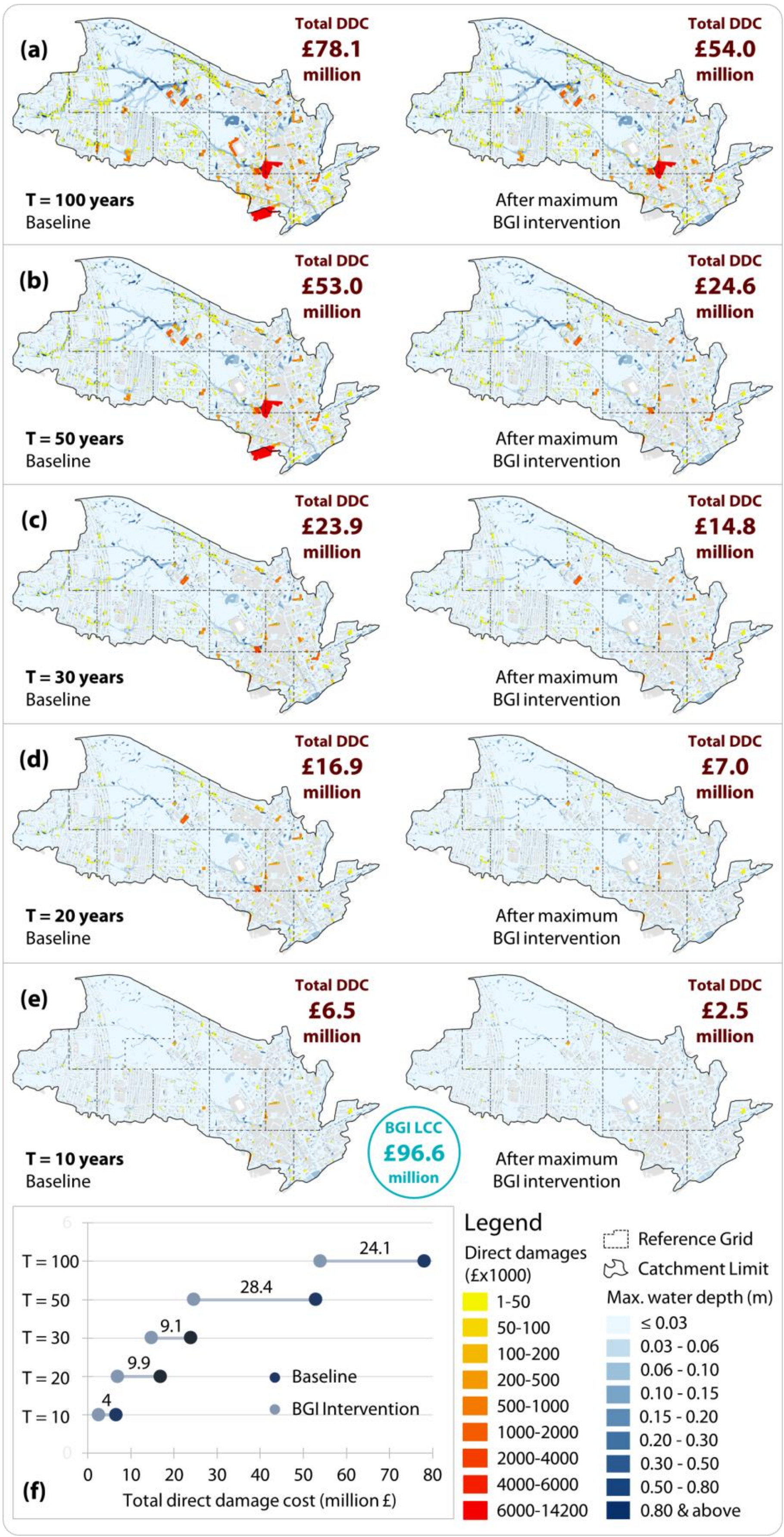

Figure 10. Maximum flood depths and direct damage cost to buildings calculated for baseline (left) and after implementing maximum BGI intervention (right) for (a) 100-year, (b) 50-year, (c) 30-year, (d) 20-year and (e) 10-year return periods, (f) shows the DDC ranges reduced by maximum intervention for considered return periods.

In terms of the cost efficiency of permeable zones by location, the zone contribution maps in Figure 9a-e show varied results. For the 100-year return period, the cost-effective zones are relatively clustered (see reference grid cells 7, 8, and 9 in Figure 9a). However, the cost-effective zones start dispersing across the catchment when moving towards lower return periods (Figure 8b-e). The cost-effective zones for the 100- and 50-year return periods share some commonalities, but the maps for shorter and longer return periods show minimal similarities.

### 3.2 Assessment of 100-year optimised Pareto front across other return periods

Figure 11a-d shows the results of the 100-year-optimised Pareto front evaluated for other return periods. The blue dots in the scatter plots represent optimised solutions for specific return periods, while the red dots indicate solutions optimised for the 100-year return period but assessed for others. The performance metric AUPF for the reference Pareto front is shown in light blue, while AUPF for the 100-year Pareto front is depicted in light blue plus reddish-filled areas. Similarly, MedRD and MaxRD are represented by blue and red dotted lines, respectively. It is clear from the scatter plots that the Pareto front optimised for the 100-year return period deteriorates when evaluated for other return periods. Aligning with earlier discussion on clustering and dispersal of cost-effective zones (Figure 9), performance metrics confirm that the 100-year Pareto front deterioration is moderate for closer return periods (30- and 50-year periods) but substantially higher for distant periods (20- and 10-year periods). Referring to Table 2, AUPFs for the 20- and 10-year periods are 132% and 153% higher than the reference AUPFs. Similarly, MedRD and MaxRD, expressed as percentages of the total risk range, are higher for the 20-year (32% and 47%) and 10-year (36% and 58%) return periods. For the 30- and 50-year periods, while MaxRD (32% and 40%) remains higher, MedRD (15% and 10%) and AUPF (53% and 73% above reference) exhibit moderate differences.

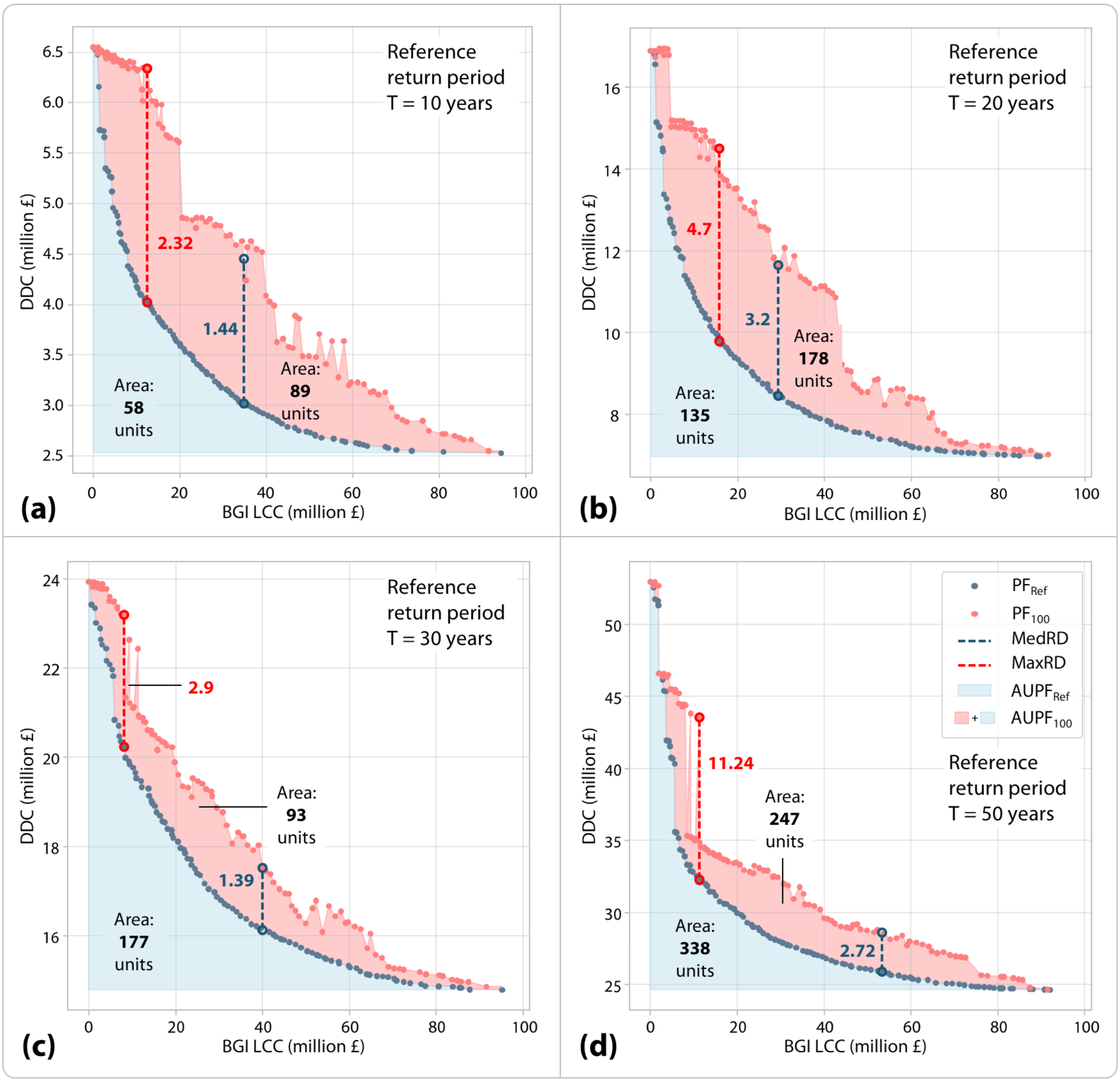

Figure 11. Solutions optimised for a 100-year rainstorm return period are assessed for (a) 50-year, (b) 30-year, (c) 20-year, and (d) 10-year rainstorm return periods. $PF_{100}$ and $PF_{Ref}$ represent solutions on Pareto fronts optimised for the 100-year and respective reference return periods. MaxRD and MedRD indicate the maximum and median differences in direct damage cost (DDC) between the 100-year and reference Pareto fronts. Similarly, $AUPF_{100}$ and $AUPF_{Ref}$ denote the area under the Pareto fronts for 100-year and reference optimisation cases.

| **Reference return period** | **$AUPF_{Ref}$** | **$AUPF_{100}$** | **$\Delta AUPF = AUPF_{100} - AUPF_{Ref}$** | **% ΔAUPF** | **Risk range (£ million)** | **MaxRD (£ million)** | **%MaxRD of risk range** | **MedRD (£ million)** | **% MedRD of risk range** |
|---|---|---|---|---|---|---|---|---|---|
| 10-year | 57.9 | 146.5 | 88.6 | 153% | 4.0 | 2.3 | 58% | 1.4 | 36% |
| 20-year | 135.3 | 313.3 | 178 | 132% | 9.9 | 4.7 | 47% | 3.2 | 32% |
| 30-year | 177.3 | 270.6 | 93.3 | 53% | 9.1 | 2.9 | 32% | 1.4 | 15% |
| 50-year | 337.8 | 584.8 | 247 | 73% | 28.4 | 11.2 | 40% | 2.7 | 10% |

Table 2. Values of performance metrics across selected return periods and their relative differences compared to the 100-year optimised Pareto front. $AUPF_{Ref}$ denotes the area under the Pareto front for the reference return period optimisation, while $AUPF_{100}$ represents the area under the Pareto front optimised for the 100-year return period and evaluated against the reference return period.

### 3.3 Optimisation for a composite of multiple return periods

The BGI optimisation results based on multiple return periods are shown in Figure 12a-c. In Figure 12a, the scatter plot displays BGI life cycle cost (LCC) along the x-axis and expected annual damage (EAD) along the y-axis, with grey dots representing generated solutions and blue dots representing optimal Pareto front solutions. Figure 12c shows the Pareto front's position on the DDC scale for the selected five return periods. EAD for each optimal solution is calculated using DDC values across all return periods, as per equations (3) and (4). Although the 100- and 50-year return periods are weighted less due to lower exceedance probabilities (3), their higher DDC values make them the largest contributors to EAD. Conversely, the 10-year return period contributes the least, as it appears only once in the EAD calculation. Figure 12b presents a spatial map of the catchment area, highlighting the contribution of individual permeable surface zones to the Pareto front solutions. An animation of the multi-return period optimisation is available in Appendix B.

Figure 12c shows that the EAD-derived Pareto front remains near-optimal on the DDC scale for most return periods, except for the 50-year one, where a section of the Pareto front is suboptimal within the 28–35 million DDC range. Unlike Pareto fronts based on individual return periods, solutions on the EAD scale (Figure 12a) remain evenly distributed due to variations in BGI performance across different rainfall intensities, balancing overall cost-effectiveness and preventing dips along the risk axis.

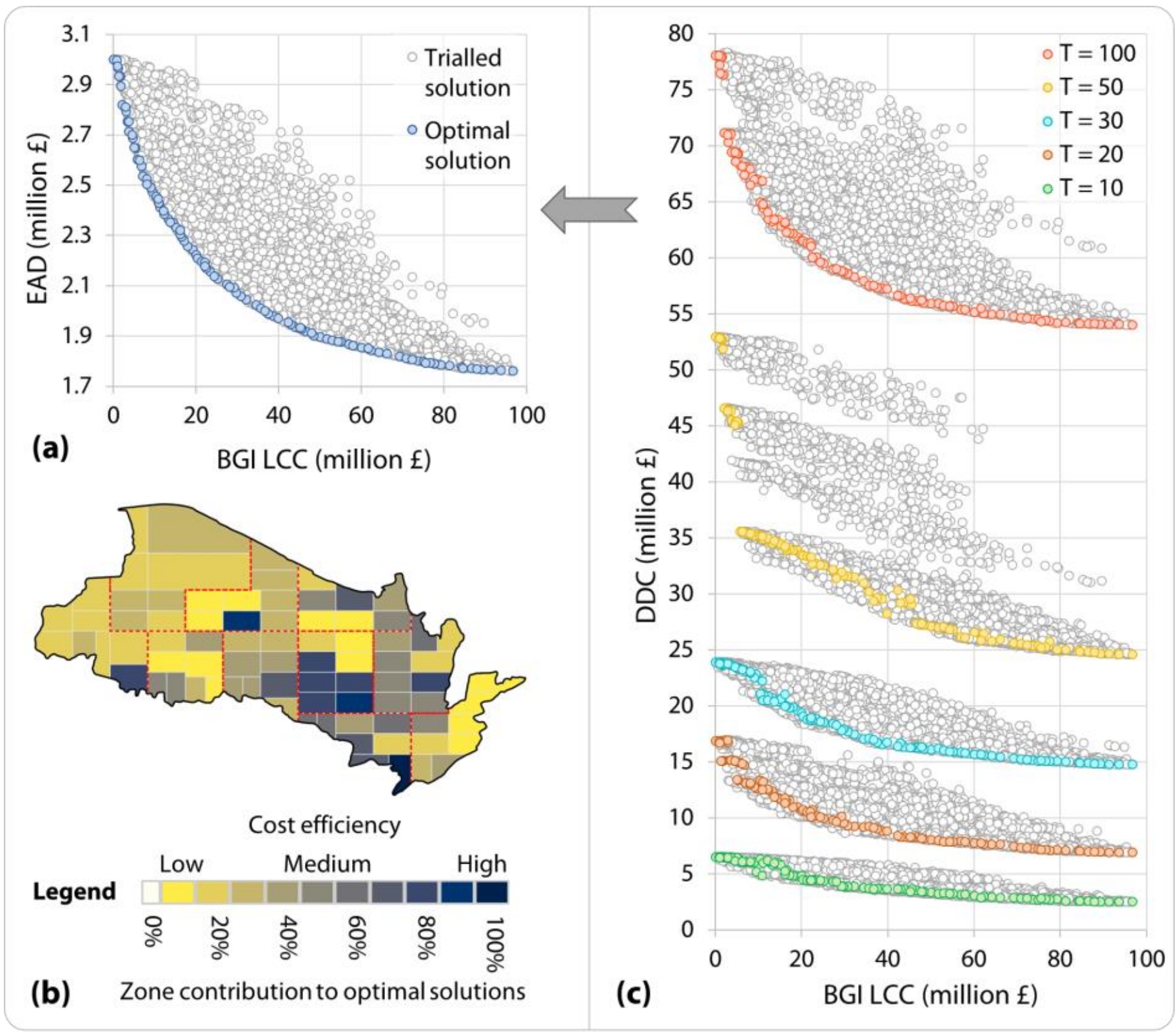

Figure 12. Multiple-return period-based BGI optimal and trialled solutions on (a) EAD and (c) DCC scales, (b) represents the contribution of permeable surface to composite optimised Pareto front. The optimisation algorithm first computes the direct damage cost (DDC) of each solution across all five return periods and then calculates the expected annual damage (EAD) from these DDC values. The optimisation is performed based on BGI LCC and EAD.

A comparison of zone contribution maps for composite optimisation (Figure 12c) and individual return period-based optimisation (Figure 9a-e) shows the composite map aligns more closely with the 100- and 50-year maps, with some similarity to the 30- and 20-year maps, and minimal overlap with the 10-year map. This pattern is expected, as BGI in higher return periods typically achieves greater DDC reductions (Figure 10 and Figure 12c), strongly influencing zone cost-effectiveness calculations.

To assess the impact of the number of return periods used to calculate EAD, *Figure S6 in Supplementary Information S6* compares EAD values calculated using five return periods (T = 10, 20, 30, 50, 100) versus three (T = 10, 30, 100), showing that fewer return periods overestimate EAD values, consistent with Ward et al. (2011).

### 3.4 Optimisation performance evaluation

The performance of the composite-optimised Pareto front in relation to the reference and the 100-year Pareto fronts is depicted in Figure 13a-d and Table 3. The visual and quantitative comparisons imply that, although the composite-optimised Pareto front does not precisely align with the reference fronts for individual return periods, it demonstrates improvements across all performance metrics and return periods compared to the 100-year-optimised Pareto front. Quantification of relative improvements in performance metrics (Figure 14) reveals the highest improvements in AUPF (73%) and MedRD (22%) and the second-highest in MaxRD (13%) for the 20-year return period. For the 10-year return period, MaxRD, MedRD, and AUPF improved by 2%, 16%, and 39%, respectively. In the 50-year return period, MaxRD improved by 23%, MedRD by just 4%, and AUPF by 26%. Conversely, the 30-year return period achieved a slightly better improvement in MedRD (7%) and a relatively smaller improvement in AUPF (15%), with no improvement in MaxRD.

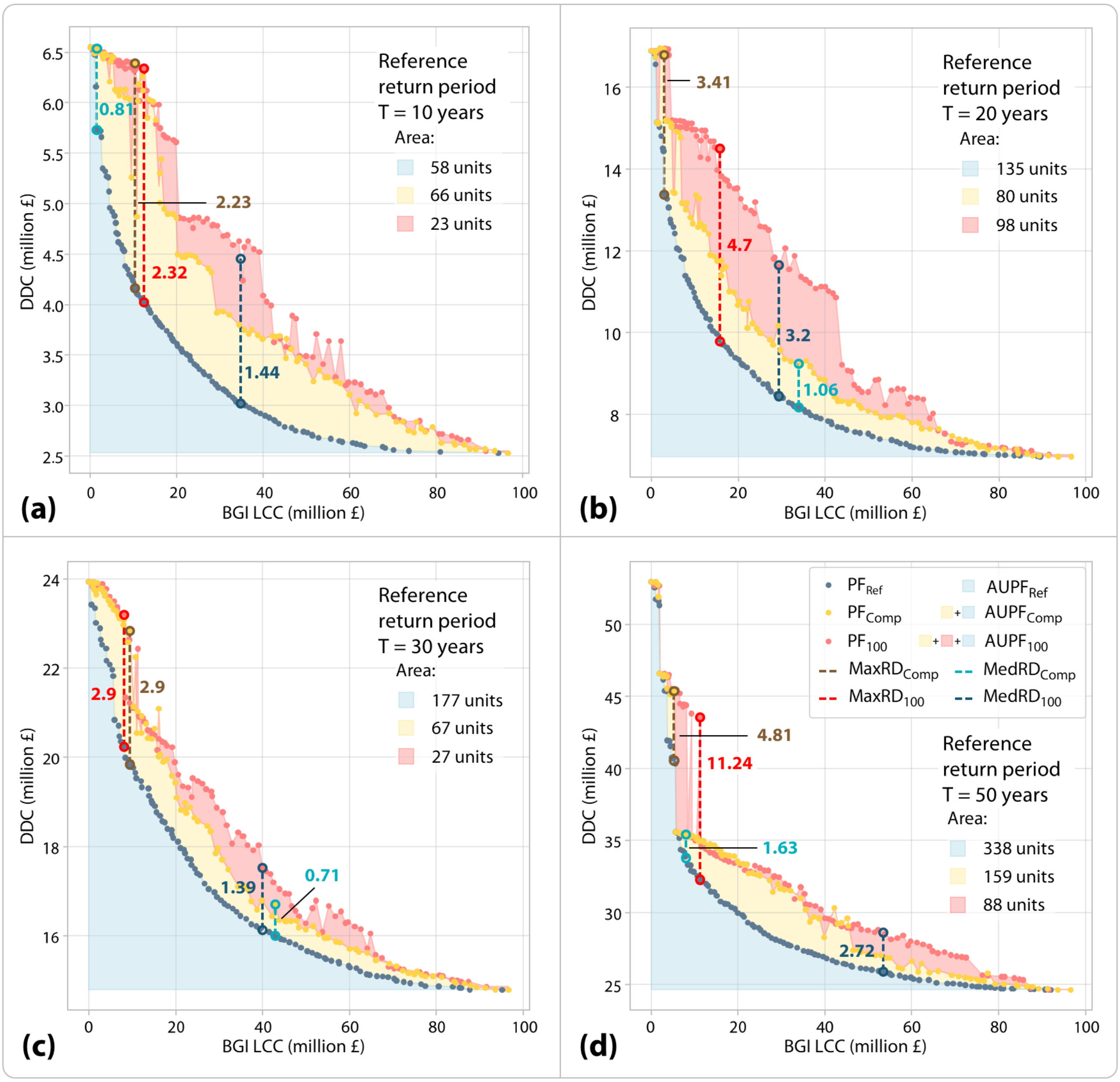

Figure 13. Solutions optimised for a 100-year and composite rainstorm return period are assessed for (a) 50-year, (b) 30-year, (c) 20-year, and (d) 10-year rainstorm return periods. $PF_{100}$, $PF_{Comp}$, and $PF_{Ref}$ represent the solutions on the Pareto fronts optimised for the 100-year, composite, and respective reference return periods. $MaxRD_{100}$ and $MaxRD_{Comp}$ indicate the maximum differences in direct damage cost (DDC) between the 100-year and reference and the composite and reference Pareto fronts, respectively. $MedRD_{100}$ and $MedRD_{Comp}$ represent the median differences in DDC between the 100-year and reference and the composite and reference Pareto fronts, respectively. Similarly, $AUPF_{100}$, $AUPF_{Comp}$, and $AUPF_{Ref}$ denote the areas under the Pareto fronts for the 100-year, composite, and reference optimisation cases.

| **Reference return period** | **$AUPF_{Ref}$** | **$AUPF_{Comp}$** | **$\Delta AUPF = AUPF_{Comp} - AUPF_{Ref}$** | **% $\Delta$AUPF** | **Risk range (£ million)** | **MaxRD (£ million)** | **%MaxRD of risk range** | **MedRD (£ million)** | **% Med RD of risk range** |
|---|---|---|---|---|---|---|---|---|---|
| 10-year | 57.9 | 124 | 66.1 | 114% | 4.0 | 2.2 | 56% | 0.8 | 20% |
| 20-year | 135.3 | 215 | 79.7 | 59% | 9.9 | 3.4 | 34% | 1.1 | 11% |
| 30-year | 177.3 | 244.1 | 66.8 | 38% | 9.1 | 2.9 | 32% | 0.7 | 8% |
| 50-year | 337.8 | 496.66 | 158.86 | 47% | 28.4 | 4.8 | 17% | 1.6 | 6% |

Table 3, Values of performance metrics across selected return periods and their relative differences compared to the composite optimised Pareto front. $AUPF_{Ref}$ denotes the area under the Pareto front for the reference return period optimisation, while $AUPF_{Comp}$ represents the area under the Pareto front optimised for the composite return periods and evaluated against the reference return period.

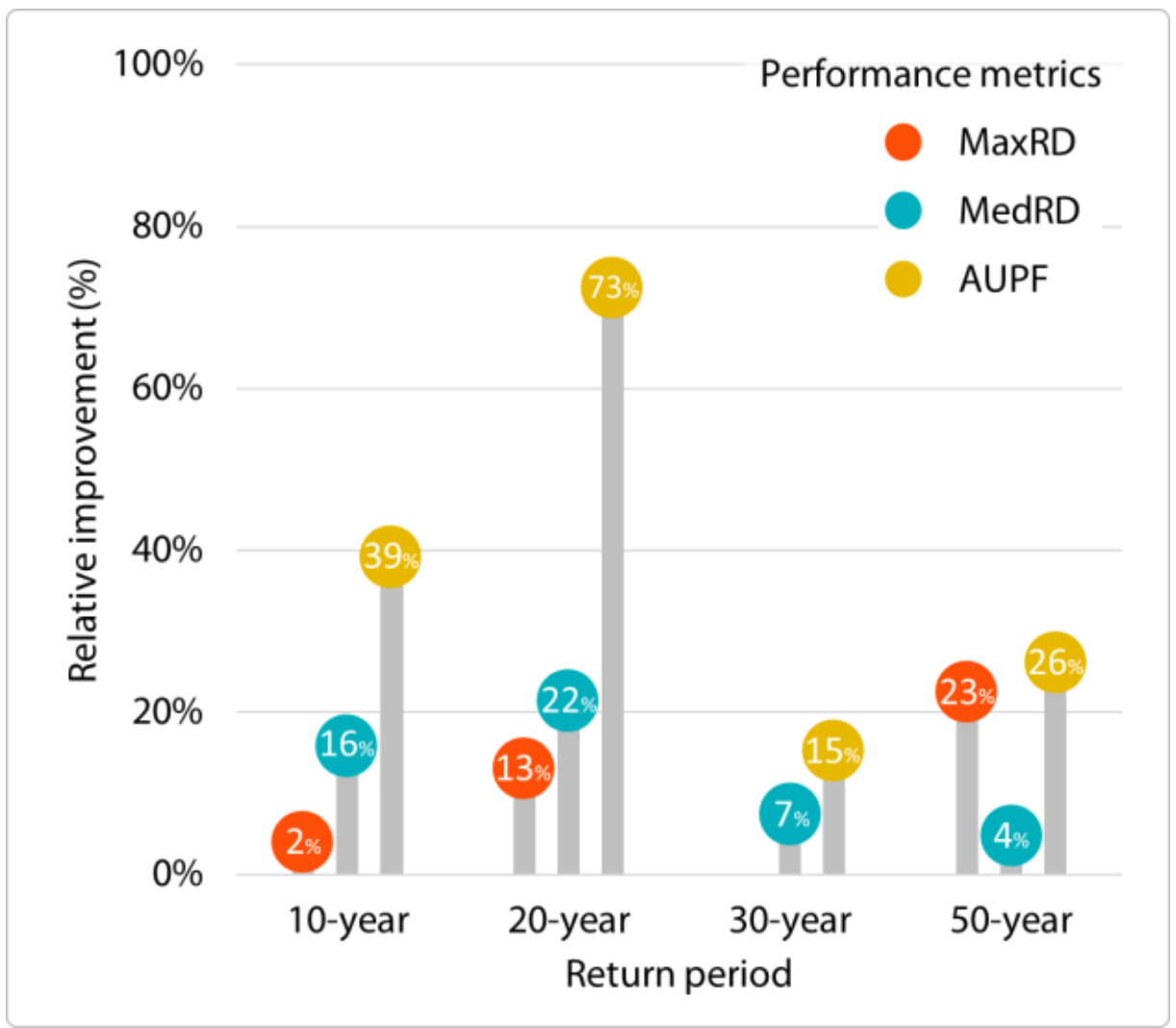

Figure 14. Improvement in performance metrics for the composite-optimised Pareto (Table 3) front compared to the 100-year-optimised front (Table 2).

### 3.5 Benefit-cost analysis

Figure 15 illustrates the benefit-cost ratios of solutions in the composite Pareto front (Figure 12a) calculated using equation (13). The figure on the x-axis represents the BGI life cycle cost (LCC) in million

£, while the y-axis shows the benefit-cost ratio. As shown in Figure 9, an increase in LCC along the x-axis indicates an increase in permeable surface zones or the overall permeable area.

The benefit-cost ratio demonstrates a sharp rise starting from the plot's baseline solution at the origin (0, 0). After reaching its maximum of 3.6 when BGI LCC is £2 million, the benefit-cost ratio gradually declines, eventually reaching a minimum of 0.5 at the maximum BGI LCC value of £96.6 million. This observation suggests that only initial investments in BGI provide considerable benefits relative to costs.

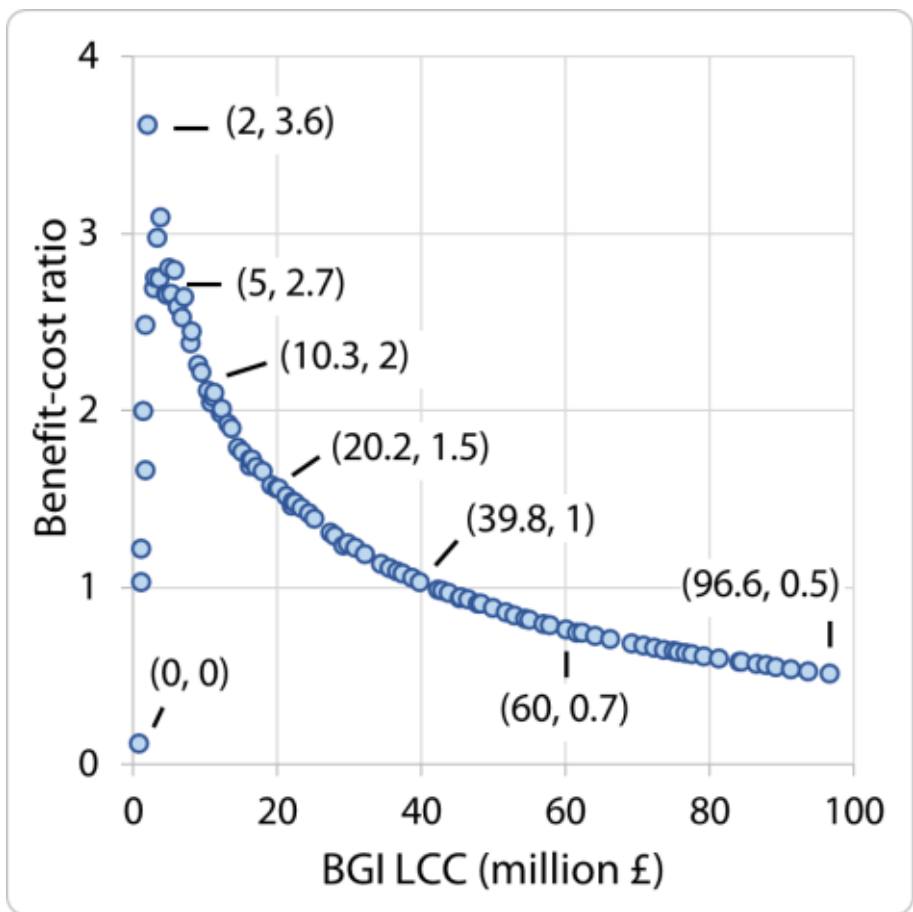

Figure 15. Scatter plots showing benefit-cost ratios for the composite optimised Pareto front.

### 3.6 Climate change resilience assessment for composite BGI design

Climate change resilience for the composite BGI design was assessed by evaluating optimal solutions under three climate uplift scenarios applied to five baseline return periods. The results, including the Pareto front with uplifts applied and the corresponding benefit-cost ratios, are presented in Figure 16b and Figure 16c. Figure 16a repeats the zone cost-efficiency map, which is the same for both composite Pareto front and uplifts. Key observations are as follows:

- **Shift in EAD values:** EAD values shift upwards with climate uplift, as higher rainfall intensities produce greater water depths around buildings, leading to increased DDC and EAD. Despite this intensification, BGI effectively reduces a broader range of EAD in climate uplift scenarios (refer to *Table S7 in Supplementary Information S7*).
- **Pareto front curvature:** The curvature of the Pareto front slightly reduces when moving from low to high climate uplift scenarios. Similar to the baseline (Figure 13a), there is a sharp reduction in EAD at the initial BGI LCC (up to ~£2 million). After this point, the rate of EAD reduction slows with increasing BGI LCC for all uplifts.

- **Benefit-cost ratio patterns:** In the benefit-cost ratio graphs (Figure 16c), initial BGI investments follow a pattern like the baseline Pareto front, yielding higher benefit-cost ratios than subsequent investments. Interestingly, the climate uplift scenarios demonstrate better benefit-cost ratios than the baseline.
- **Shape of composite Pareto front:** Although a few gaps in the climate uplift fronts indicate an uneven distribution of some solutions, they remain intact overall, following the curvature line.

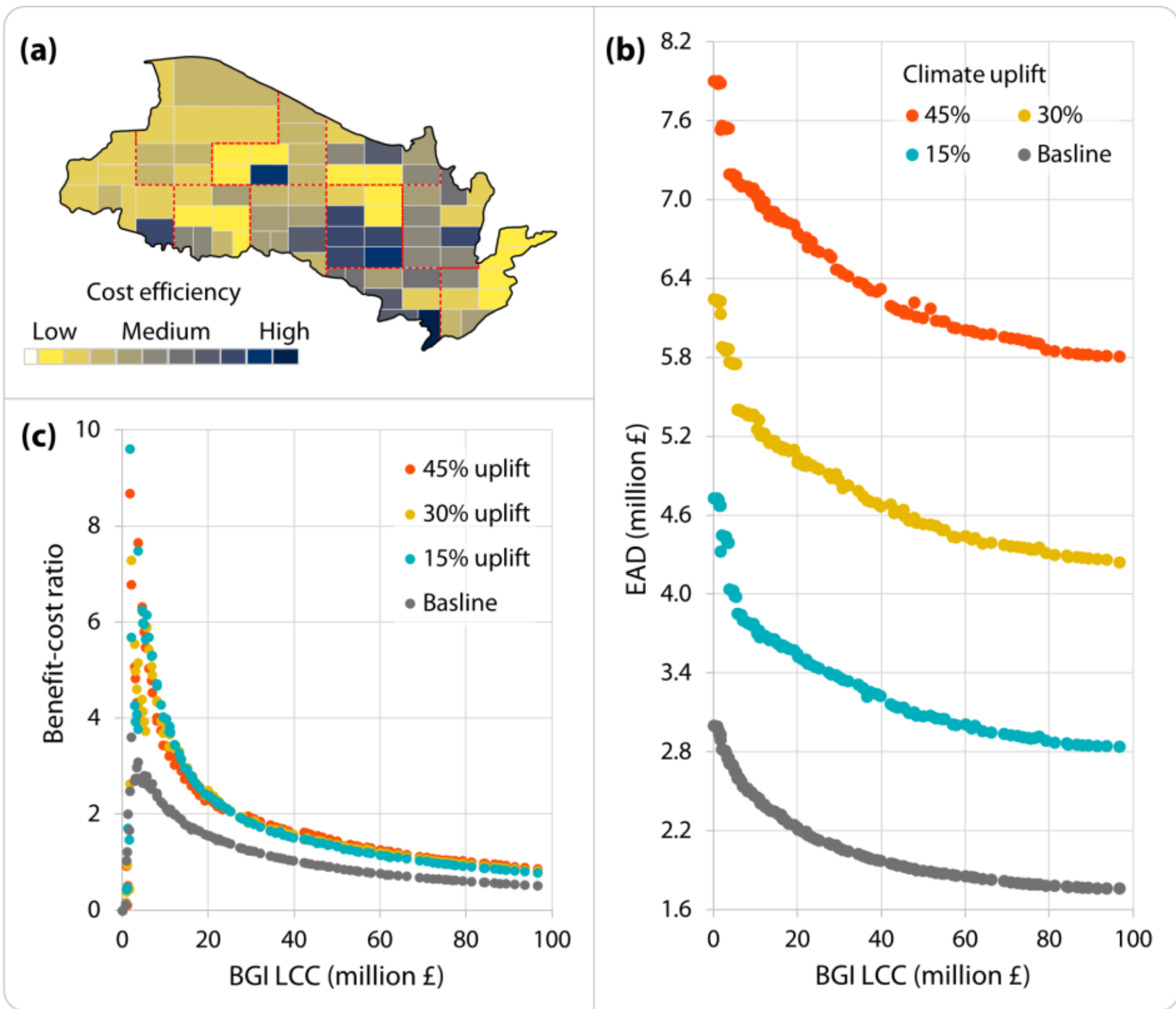

Figure 16. Composite optimisation-based (a) zone efficiency map, (b) Pareto front assessment for climate change uplifts, and (c) benefit-cost ratios for the uplifts.

## 4 Discussion

This study examines how varying rainstorm intensities affect optimised BGI designs and evaluates a composite optimisation approach aimed at improving the universality of a single FRM design. This section discusses the underlying causes and practical implications of the results presented in Section 3.

### 4.1 Influence of rainstorm magnitude and risk objective function on BGI performance

The results demonstrate the variation in curvature (minimisation) of Pareto fronts optimised for distinct rainstorm return periods. This variation is likely due to the catchment's hydrodynamics across different return periods and how the risk function operates. Different rainfall intensities result in varying water depths around buildings (Iliadis et al., 2023), and the effectiveness of permeable intervention zones depends on the generated surface runoff and flow pathways. For the 50-year return period, a smaller fraction of the total available permeable area effectively reduces water depths around buildings, leading to a substantial decrease in DDC. Another factor could be associated with the formulation of the risk objective, which considers the type of buildings exposed during each return period. As shown in Figure 2, non-residential buildings typically have larger footprint areas and are more likely to be exposed to a higher DDC. It is evident from Figure 10 that more non-residential buildings are at risk during the 50- and 100-year return periods. Due to higher risk, individual or combinations of a few permeable surface zones (covering a smaller area) at the start of the Pareto front appear to reduce higher DDC more effectively for the 50-year return period, as suggested by Figure 10a-b.

Another notable observation in the distinct Pareto fronts is the gaps between successive optimal solutions (see the blue line arrows along the y-axes in Figure 9a-d). These gaps can be attributed to the non-linearity of the risk function, as explained below:

i. BGI interventions reduce the water depth around a building. If the water depth does not meet the ‘not at risk’ criteria outlined in equation (2), the building remains at risk but with a reduced DDC.
ii. If the BGI intervention reduces the water depth to a level that meets the ‘not at risk’ criteria, the DDC for that specific building becomes zero.

The first type of DDC reduction, while the building is still at risk, is more gradual. However, the second situation can cause an abrupt decrease in total DDC. This observation is particularly true for non-residential buildings, where a minimal intervention (small LCC) that slightly reduces water depth to meet ‘not at-risk criteria’ can prevent a large building from being at risk, leading to a higher DDC reduction and the gaps between successive solutions on the Pareto fronts. For further clarification, please refer to *Figure S5 Supplementary Information S5*. A similar discontinuity along the Pareto front has also been observed by Hou & Yuan (2020), albeit using a different risk objective function within the SUSTAIN model framework. These gaps underscore the critical importance of optimisation-based BGI design in identifying scenarios where a marginal increase in intervention size, or associated LCC, can yield disproportionately higher benefits, making such solutions highly cost-effective and strategically superior for decision-makers.

The clustering and dispersal of cost-efficient permeable zones (Figure 9a-e) from longer to shorter return periods can again be attributed to catchment hydrodynamics and risk function working. During high-intensity rainstorm events, such as the 100-year return period, water flows across the different regions of the catchment from the northwest to the southeast (Figure 10a). These regional flow paths combined with local surface runoffs potentially place many non-residential buildings in reference grid cells 7, 8, and 9 at risk. Consequently, permeable zones in these grids are not only expected to infiltrate local surface runoff but also maximally intercept flow paths from the upper parts of the catchment, reducing water depth around non-residential buildings and making a cluster of cost-effective zones in this region. For lower-intensity rainstorms, such as the 10-year return period (Figure 10e) with probably no considerable long flow paths from the upper catchment, risk to the buildings is predominantly influenced by local surface runoffs, resulting in scattered cost-effective zones. Other return periods between the 100- and 10-year follow a similar mechanism with varying contributions from local and regional runoffs.

Quantitative analysis shows that the BGI Pareto front optimised for a 100-year rainstorm did not align well when assessed against other considered return periods. A careful examination of Figure 9 and Figure 11 reveals that the initial solutions on Pareto fronts include either a single best zone or combinations of a few top-performing zones. As LLC increases, additional zones, ranging from good to less effective, are incorporated. Since the most cost-effective zones and combinations in the 100-year optimisation differ from others, they perform poorly, especially for shorter return periods (10 and 20 years). However, discrepancies are relatively smaller at the Pareto fronts' tail, where reference and 100-year fronts already include less effective zones to maximise intervention. These results indicate that the effectiveness of permeable surface zone locations and combinations is highly sensitive to rainfall intensity. This sensitivity is potentially linked to varying infiltration rates under different surface runoff velocities and associated soil saturation (Huang et al., 2013; Reeder et al., 1980), differing water depths around buildings, and the non-linear response of depth-damage curves (Figure 5), demonstrating that BGI optimisation based on single maximum return periods fails to deliver robust solutions for storm variations.

The lack of generalisation in distinct return period-based optimised BGI designs is consistent with the general findings of Chen et al. (2024) and the more specific results reported by Ur Rehman et al. (2024), though limited to evaluations of the 100- and 30-year return periods. However, in this study, the comparison of the 100-year return period Pareto front assessed for the 30-year return period shows a relatively low discrepancy, possibly due to the adaptation of the continuous risk objective function. When comparing cost efficiency by zone, spatial patterns differ between the studies due to variations in the risk objective function. The reference study, using a discrete risk function based on the count of buildings exposed to flooding, gives equal weight to all buildings, favouring cost-effective zones in residential areas. In contrast, the current study uses a continuous risk function based on DDC, shifting cost-effective zones to non-residential areas due to larger building footprints and associated damage costs. Additionally, the continuous risk function in this study effectively captures even minor flood reductions, resulting in fewer 'no contribution' zones compared to a larger number of 'no contribution' zones in the reference study. The comparative results emphasise the importance of carefully selecting a risk objective function as it can significantly influence the cost-effective locations for BGI interventions. The efficiency and reliability of DDC in capturing risk across different building types make it an effective risk objective function for optimisation. Although the current study focuses solely on buildings as flood receptors, the use of DDC can also allow for the integration of other receptors, such as transport routes (Martínez-Gomariz et al., 2019; Pregnolato et al., 2017), and critical infrastructure and services (Murdock et al., 2018), by converting their potential risk into monetary values, enabling the formulation of a compound risk objective function for comprehensive BGI performance optimisation.

DDC as a risk objective is highly influenced by building type, often favouring non-residential areas in BGI optimisation. While DDC, calculated for residential and non-residential buildings differently, remains the standard method for assessing flood damages to buildings in the UK, its reliability depends on accurate building classification. Misclassified data can distort DDC estimates and spatial patterns and misguide BGI prioritisation. Another factor that can impact DDC values is the accuracy of flood depths produced by the hydrodynamic model, which primarily depends on the spatial resolution of the Digital Elevation Model (DEM). Although very high-resolution LiDAR DEM is generally preferred, it is often prone to outliers due to complex processing and significantly increases computational cost (Saksena, 2015). Therefore, a balance between DEM resolution and hydrodynamic computational efficiency is essential. The current study achieves this balance by using a 5-metre DEM.

### 4.2 Effectiveness of multi-return period optimisation for BGI design

Multi-return period optimised BGI design generally exhibited good compliance across all the considered return periods, overcoming the incompatibility of single-return period-based design. Reflecting on the composite-optimised solutions, achieving their simultaneous alignment with all four Pareto fronts optimised for individual return periods is nearly impossible. The goal is to find the best trade-offs that minimise disparities across all return periods, thereby introducing robustness into the BGI design. Composite optimisation achieves this by proportionally considering risk reductions from each return period during EAD calculation. As shown in Figure 12c, a few solutions are slightly suboptimal for the 100-year return period on the DDC scale. However, this trade-off is beneficial, as minimal degradation in the 100-year performance leads to significant improvements in BGI efficiency for other return periods. Figure 13 illustrates that longer return periods generally exhibit larger DDC ranges and greater maximum and median DDC differences than shorter ones. Nevertheless, low- to moderate-intensity rainstorms, which occur more frequently, make it critical to consider BGI performance across shorter return periods. Adopting EAD as a risk objective function effectively addresses this balance, though longer return periods still exert a strong influence as BGI reduces higher DDC in those. Additionally, the use of $D(S_Z)_{INFIN}$ in equation (4) still maintains a significant focus on the 100-year return period while facilitating considerable improvements in other return periods. For instance, despite contributing only once during EAD calculation in equation (3), solutions still see improvement on the 10-year return period scale. Overall, a BGI design based on composite optimisation is more robust than one based on a 100-year (single maximum) return period. The variation in the performance of the proposed design across return periods is influenced by catchment hydrodynamics and the non-linear effectiveness of BGI in reducing damage across different rainfall intensities, as discussed in Section 4.1.

Iliadis et al. (2024) employed a limited scenario-based source-receptor framework to prioritise different BGI interventions based on their economic viability, considering DDC for each return period separately.

However, such comparisons, without standardising DDCs, may lead to misinterpretation. In contrast, an adaptation of EAD in the current study standardises damages by considering the relative contribution from each return period, providing a single, consistent metric for BGI performance evaluation. The use of EAD is, therefore, not only instrumental for optimisation but also critically important for determining the economic viability of different BGI solutions at the same single scale, ultimately enhancing the decision-making process for BGI selection.

As EAD is derived from multiple DDC values, it inherits all the sensitivities and uncertainties associated with DDC, as discussed in Section 4.1. An additional source of uncertainty is the non-linearity of DDCs across different return periods, which complicates determining the appropriate number of storm return periods to include in the calculation. As supported by Ward et al. (2011) and illustrated in *Figure S6 of Supplementary Information S6*, using only a few return periods (e.g. three DDCs) can lead to an overestimation of EAD. However, including many return periods is often impractical due to the computational demands of detailed hydrodynamic flood modelling. This study finds a balance by using five return periods, offering a reasonable trade-off between computational efficiency and EAD accuracy.

Despite accounting for exceedance probabilities to ensure proportional contributions from each return period, zone contribution maps reveal that efficient permeable surface zones still tend to be skewed towards higher return periods, primarily due to the larger DDC reductions associated with these rarer but more severe events. Nonetheless, this skewness reflects a balanced outcome inherent to the standard EAD calculation method. When more alignment with shorter return periods is desirable, higher weighting can be applied to frequent flood events in the EAD calculation during optimisation. However, cost-benefit analysis should still be based on the standard EAD calculation method.

### 4.3 Economic viability and resilience of composite optimised BGI solutions

The benefit-cost analysis (Figure 15) indicates that a couple of initial solutions (after baseline), which involve minimal zone combinations (e.g., only one or two zones), have relatively low benefit-cost ratios due to their limited impact on damage reduction. However, as additional zones are introduced, the optimisation algorithm identifies good combinations that significantly reduce damage costs. This substantial reduction is likely due to some large, high-risk, non-residential buildings transitioning from “at-risk” to “not-at-risk,” thereby boosting the benefit-cost ratio. As the BGI LCC increases beyond the point of maximum benefit-cost, the values of the benefit-cost gradually decline, indicating that additional permeable zones do not proportionally reduce the damages to the buildings. As expected, the results show that only strategically located permeable zones, highlighted as darker blue areas in Figure 12b, are cost-effective in reducing risk. This finding underscores the value of a fully distributed 2D model, which preserves the spatial fidelity of permeable areas within each zone and enables precise identification of optimal intervention locations, unlike semi-distributed models that generalise interventions at the sub-catchment level (Chen et al., 2024; Huang et al., 2022). The results also highlight the advantage of an automated optimisation framework for 2D modelling over scenario-based approaches, such as the one presented by Iliadis et al. (2024), which relies on a limited set of broad intervention scenarios and may result in suboptimal designs. In contrast, automated algorithms efficiently evaluate a large number of fine-grained intervention options to intelligently identify precise locations and combinations for BGI implementation.

The lifetime benefit-cost ratios of permeable surface interventions remained low, consistent with the findings of Iliadis et al. (2024). Catchment-specific factors contributing to this low return include a relatively steep catchment gradient and low soil permeability in the study area. In steeper catchments, water flows rapidly, leaving less time for infiltration through permeable surfaces (Lucke & Beecham, 2013). Similarly, low soil permeability limits the volume of water that can infiltrate (Holman-Dodds et al., 2003). Due to catchment-specific characteristics, the findings on the economic viability of permeable surface interventions are not generalisable. Further studies are needed across diverse catchment types and rainstorm durations. Additionally, these estimated BGI benefits in this study are based solely on reductions in direct damage costs to buildings. Including other costs—such as traffic and emergency service disruptions, business interruption, and environmental benefits—would likely increase the overall benefits.

Irrespective of catchment characteristics, climate uplift stress testing (Figure 16) demonstrated the robustness of the proposed composite-optimised BGI design for increasing severity of rainstorms. The observed gaps in the Pareto fronts (Figure 16b) under the climate uplift scenario indicate highly cost-effective solutions, where a slight increase in BGI LCC yields substantial reductions in EAD. This may result from non-residential buildings shifting from at-risk to not-at-risk categories due to flood depth reductions. The improved benefit-cost ratios (Figure 16c) under climate uplift scenarios can be attributed to higher DDCs and the resulting increase in EAD, which enables BGI to achieve greater damage reductions.

In a broader context, these findings on cost-benefits and robustness of BGI design should be interpreted with caution. The estimation of lifetime benefit-cost ratios assumes that catchment conditions, such as land use, flood receptors and soil permeability, remain unchanged over the lifespan of BGI interventions. However, these features are likely to evolve over time (Butters et al., 2025), which could influence the optimisation results and related benefit-cost ratios. Nevertheless, future changes in urban infrastructure are generally expected to be gradual and more directly controllable by humans, whereas climate change can cause more abrupt shifts with minimal human control (M. Chen et al., 2021). Therefore, despite these limitations, the findings of this study remain highly relevant and important for urban flood risk management planning.

## 5 Conclusions

This study systematically evaluated different optimisation approaches across five storm magnitudes (T = 10, 20, 30, 50, and 100 years) to identify a cost-effective and robust Blue-Green Infrastructure (BGI) design for urban flood risk management. By integrating a detailed flood model with the Non-dominated Sorting Genetic Algorithm II (NSGA-II), optimal locations and combined sizes for BGI interventions were identified by separately minimising Life Cycle Cost (LCC) against Direct Damage Cost (DDC) and Expected Annual Damage (EAD), offering new insights and practical strategies for informed, resilient urban flood risk management. The findings support the following key conclusions:

- Adopting a continuous risk objective function, such as DDC, is vital to capture an accurate spatial distribution of risk and to reflect the full variability of BGI performance. DDC not only accounts for different building types but is also instrumental in developing a composite risk function for multiple rainstorms, such as EAD, which effectively translates a complex many-objective optimisation problem into a manageable multi-objective one for easy implementation and adaptation.

- The efficiency of permeable surface locations and their combinations is highly sensitive to rainstorm intensity, potentially driven by variations in surface runoff, time-dependent infiltration rates, and spatial risk distribution. The study shows that these sensitivities result in poor transferability of the 100-year optimised design to other return periods. Therefore, relying solely on designs optimised for a single-maximum return period can lead to severe performance degradation during other events, particularly more frequent, lower-magnitude storms. Consequently, implementing BGI based on such designs is highly prone to compromising long-term returns on investment.
- Optimisation using EAD, calculated by integrating the proportional contributions of DDCs from multiple rainstorm return periods, achieves a balanced trade-off across storm magnitudes, effectively managing associated sensitivities and significantly enhancing the adaptability and effectiveness of BGI design. The composite design performs effectively across a range of storm intensities—from severe events to moderate and lighter storms—thereby strengthening urban flood resilience under climate change and maximising lifetime return on investment. This key finding recommends a paradigm shift from traditional single-return-period approaches towards composite return period-based urban flood risk management.
- Lifetime cost-benefit analysis, enabled by LCC and EAD, is critical to rank the economic viability of optimised BGI interventions. The study revealed that benefit-cost ratios for optimised permeable surface interventions exhibit a non-linear relationship with BGI cost, rising to a point of maximum benefit before declining, underscoring the importance of identifying economically efficient designs. Such analysis can significantly aid practitioners in making more informed decisions for BGI implementation.

Despite these contributions, several limitations must be acknowledged. Simulating multiple storms in a fully distributed hydrodynamic model to formulate a composite risk function requires considerable computational resources. While reducing the number of return periods could mitigate computational burden, accurately calculating EAD still requires adequately sampled storm events due to the non-linearity of DDC. Users need to maintain a balance between computational cost and design reliability. In addition to these modelling challenges, key factors that can influence design performance include catchment slope gradient, soil permeability, storm duration, and changes in land use or dwellings over time. To advance the work presented in this study, future research should focus on incorporating additional BGI features—such as detention ponds, swales, and green roofs—alongside longer storm durations and a compound risk function that also accounts for critical infrastructure, emergency services, and transport networks. These steps are expected to further strengthen BGI appraisal and improve the economic feasibility of urban flood management strategies.

**Funding**

The Natural Environmental Research Council (NERC) funded this research through the ONE Planet Doctoral Training Partnership (Grant: NE/S007512/1).

**Conflict of interest**

The authors declare no conflict of interest.

**Appendix A: Supplementary information**

**Appendix B: Animation of the multi-return period-based BGI optimisation process**